\documentclass[12pt]{article}
\usepackage{fullpage}
\usepackage{times}
\usepackage{epsfig}
\usepackage{graphicx}
\usepackage{grffile}
\usepackage{amsmath}
\usepackage{amssymb}
\usepackage{color}
\usepackage{bbm}
\usepackage{comment}
\usepackage{subcaption}
\usepackage{array}
\usepackage[ruled,vlined]{algorithm2e}
\newcommand{\bx}{\mathbf{x}}
\newcommand{\bz}{\mathbf{z}}

\newcommand{\bn}{\mathbf{n}}
\newcommand{\bc}{\mathbf{c}}
\newcommand{\ba}{\mathbf{a}}

\newcommand{\bv}{\mathbf{v}}

\newcommand{\commentPP}[1]{\textcolor{blue}{\hspace{0.2in} {\bf  PP: } {#1}}\hspace{0.2in}}

\newcommand{\commentBlock}[1]{}

\title{Towards causal benchmarking of bias \\ in face analysis algorithms}

\author{G. Balakrishnan$^{\dag\ddag}$ \and Y. Xiong$^\ddag$  \and W. Xia$^\ddag$ \and P. Perona$^{*\ddag}$ }
\date{ 
$\dag$ Massachusetts Institute of Technology \\
$*$ California Institute of Technology \\ 
$\ddag$ Amazon Web Services\\
}

\begin{document}
\maketitle

\begin{abstract}
Measuring algorithmic bias is crucial both to assess algorithmic fairness, and to guide the improvement of algorithms. Current methods to measure algorithmic bias in computer vision, which are based on {\em observational} datasets, are inadequate for this task because they conflate algorithmic bias with dataset bias.

To address this problem we develop an {\em experimental} method for measuring algorithmic bias of face analysis algorithms, which manipulates directly the attributes of interest, e.g., gender and skin tone, in order to reveal causal links between attribute variation and performance change. Our proposed method is based on generating synthetic ``transects'' of matched sample images that are designed to differ along specific attributes while leaving other attributes constant. A crucial aspect of our approach is relying on the perception of human observers, both to guide manipulations, and to measure algorithmic bias.

Besides allowing the measurement of algorithmic bias, synthetic transects have other advantages with respect to observational datasets: they sample attributes more evenly, allowing for more straightforward bias analysis on minority and intersectional groups, they enable prediction of bias in new scenarios, they greatly reduce ethical and legal challenges, and they are economical and fast to obtain, helping make bias testing affordable and widely available. 

We validate our method by comparing it to a study that employs the traditional observational method for analyzing bias in gender classification algorithms. The two methods reach different conclusions. While the observational method reports gender and skin color biases, the experimental method reveals biases due to gender, hair length, age, and facial hair.

\end{abstract}


\newpage

\section{Introduction}
Automated systems trained using machine learning methods are increasingly used to support decisions in industry, medicine and government. While performance of such systems is often excellent, accuracy is not guaranteed, and needs to be assessed through careful measurements. Measuring {\em biases}, i.e., performance differences, across protected attributes such as age, sex, gender, and ethnicity, is particularly important for decisions that may affect peoples' lives. Unlike systems based on human judgment, where measuring and correcting biases is notoriously difficult, measuring and mitigating algorithmic bias is feasible and may become a powerful agent of progress towards more fair, accountable and transparent institutions~\cite{kleinberg2019,Mullainathan2019}.

\begin{figure}[t]
\centering
\includegraphics[width=\textwidth]{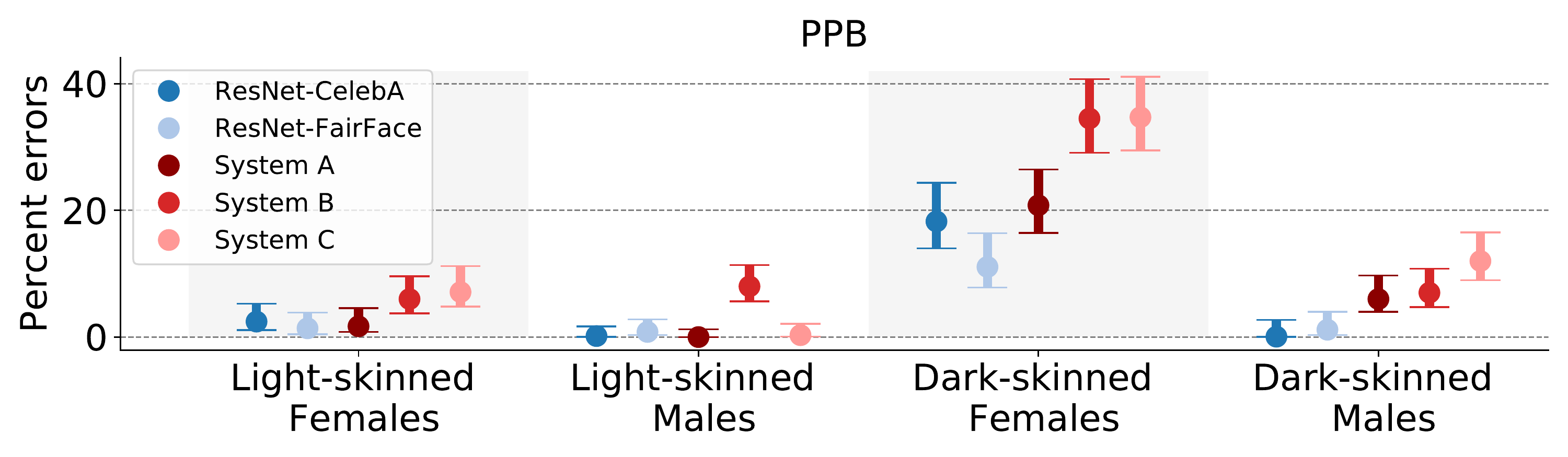}\\
\includegraphics[width=\textwidth]{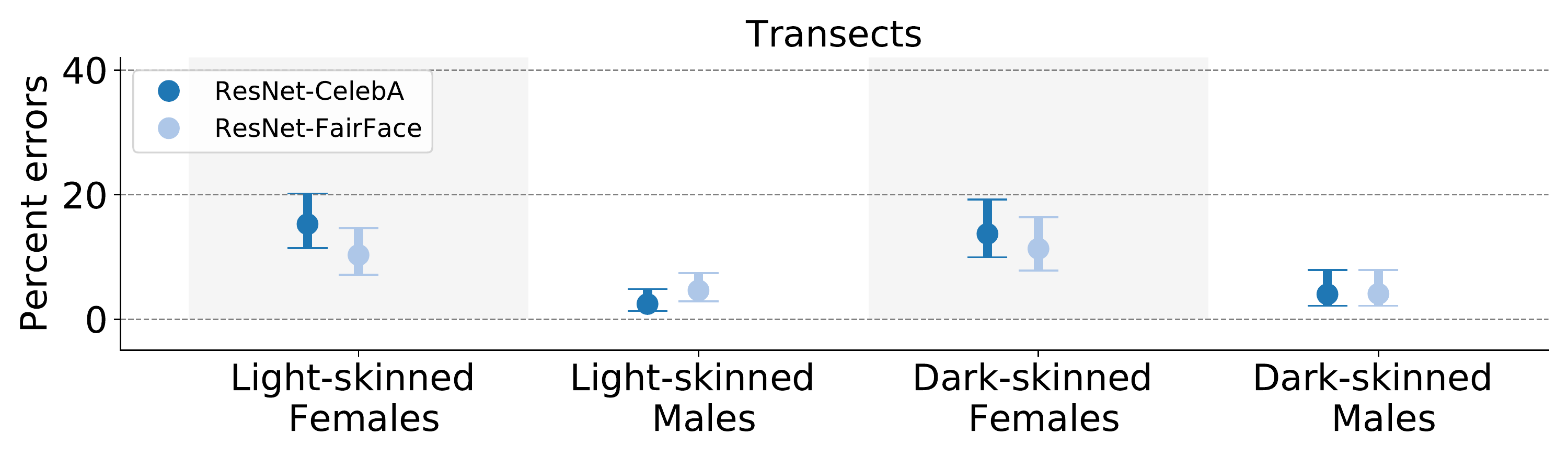}
\caption{\footnotesize{\bf Algorithmic bias measurements are test set dependent}. (Top) Gender classification error rates of three commercial face analysis systems (System A--C) were measured in 2017 on the Pilot Parliaments Benchmark (PPB)~\cite{buolamwini2018gender}, an observational dataset of portrait pictures downloaded from the web sites of six national parliaments in Scandinavia and Africa. Error rates for dark-skinned females were found to be significantly higher than for other groups. We observed the same qualitative behavior when we replicated the study by training a standard classifier (ResNet-50) on two publicly available face datasets (CelebA, FairFace) and testing the two models thus obtained on a replica of the PPB dataset. (Bottom) Our experimental investigation using the Transects dataset, where sample faces are matched across attributes, reveals a different picture of algorithmic bias (see Fig~\ref{fig:bar-plots}, Sec.~\ref{sec-experiments}, and~\ref{sec-discussion-conclusions} for a more complete analysis).}
\label{fig:gender-shades-plots}
\end{figure}

The prevailing technique for measuring the performance of algorithms is to measure statistics like error frequencies on a test set that is sampled {\em in the wild}, hopefully mirroring some of the data statistics that will be encountered in the field.
Studies of algorithmic bias in computer vision~\cite{buolamwini2018gender,brandao2019age, klare2012face, Krishnapriya2019} have adapted this approach by adding one additional step: each image of the test set is annotated for attributes of interest (e.g., ethnicity, gender and age), and the test set is then split into groups that have homogeneous attribute values. Comparing error rates across such groups yields predictions of bias. As an example, Fig.~\ref{fig:gender-shades-plots}-top shows the results of a recent study of algorithmic bias in gender classification of face images. This type of study is called {\it observational}, because the independent variables (e.g., skin color and gender) are sampled from the environment, rather than controlled by the investigator.
  
Algorithmic bias is measured for two reasons. First, fairness: would changing a protected attribute, all else being equal, cause a systematic change in the output of the algorithm? For example, would two job applicants, that differed only by their gender or ethnicity, face predictably different outcomes~\cite{bertrand2004emily}?
The second reason for measuring bias is getting rid of it: which actions should one take to best improve the system's performance? For example, should the engineers who are in charge of developing systems A, B, and C (Fig.~\ref{fig:gender-shades-plots}, top) infer that the best strategy is to add more examples of dark-skinned women to their training set? Thus, measuring algorithmic bias ultimately has one goal: revealing causal connections between attributes of interest and algorithmic performance.

Unfortunately, observational studies are ill-suited for drawing such conclusions. When one samples data in the wild, other variables may correlate with the variable of interest, and any one of the correlated variables may have an influence on the performance of the algorithm. 
Thus, it is difficult to impute the cause of performance differences to variations in the variable of interest -- as the old saying goes: {\em ``correlation does not imply causation.''} 

One simple instance of this problem is sample bias: samples in the wild may fail to represent specific combinations of variables of interest~\cite{kearns2017preventing,kearns2019empirical,kearns2019ethical}. For example, the appearance of the parliamentarians in the PPB dataset~\cite{buolamwini2018gender} tends to be gender-stereotypical, e.g., very few males have long hair and almost no light-skinned females have short hair (Fig.~\ref{fig:violin-color}, and~\cite{muthukumar2018understanding}). The fact that hair length (a variable that may affect gender classification accuracy) is correlated in PPB with skin color (a variable of interest) complicates the analysis. In addition, the sample dataset that is used to measure bias is often not representative of the population of interest. For example, the middle-aged Scandinavians and Africans of PPB are not representative of, say, the broad U.S. Caucasian and African-American population~\cite{lohr2018}. While observational methods do yield useful information on disparate impact within a given test set population, generalizing observational performance predictions to different target populations is hit-or-miss~\cite{torralba2011unbiased} and can negatively impact underrepresented, or minority populations~\cite{merkatz1993women,simon2005wanted}. In a nutshell, one would want a method that systematically identifies algorithmic bias while transcending the peculiarities of specific test sets.

Scientists in biology, medicine and the social sciences are well aware of this problem and have developed practices to discover, and to control for, confounding variables. A powerful approach to discovering cause-effect relationships is the {\em experimental method} which involves artificially manipulating the variable of interest, while fixing all the other inputs~\cite{bertrand2004emily,pearl2009causality}. This is not easy in the case of image data, leading us to ask the question: {\em Can one systematically measure bias in computer vision algorithms using the experimental method?} While this is not immediately intuitive~\cite{muthukumar2018understanding}, we find that the answer is yes, and offer a practical way forward.

\begin{figure}[t!]
\vspace{-0.2in}
\includegraphics[width=\textwidth]{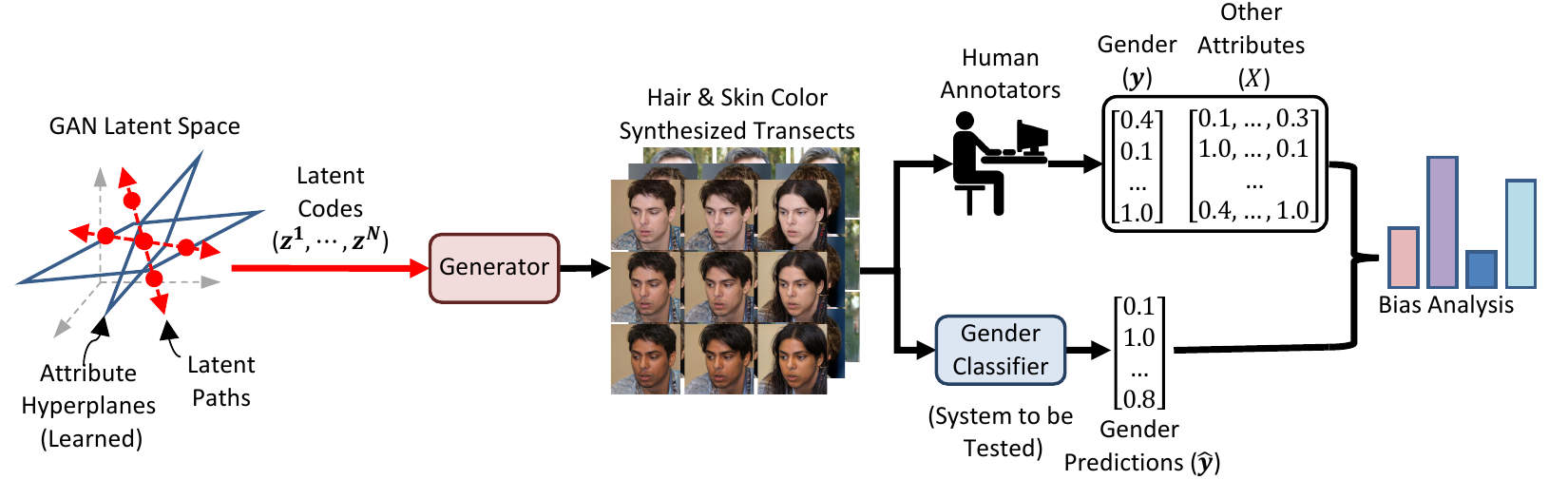}
\caption{\footnotesize{\bf Synopsis of our approach.} A generative adversarial network (Generator) is used to synthesize  ``transects,'' or grids of images, modifying selected attributes on synthetic faces (in this example: hair length and skin tone). This is accomplished by traversing the generator's latent space in attribute-specific directions. These directions are learned using randomly sampled faces and human annotators (not shown). Human annotations on the transects provide generator-independent ground truth to be compared with algorithm output to measure algorithm errors. Attribute-specific bias measurements are obtained by comparing the algorithm's predictions with human annotations as the attributes are varied. The depicted example may study the question: {\em Does hair length, skin tone, or any combination of the two have a causal effect on classifier errors?} Transects exploring other attributes are shown in Fig.~\ref{fig:single-transect-examples},~\ref{fig:multi-transect-examples}, and~\ref{fig:octets}(a). The GUIs for human image annotation are shown in Fig.~\ref{fig:screenshots}. Samples of image annotations are shown in Fig.~\ref{fig:annotated-long-transects}.}
\label{fig:teaser}
\end{figure}

Our approach (Fig.~\ref{fig:teaser}) generates the test images synthetically, rather than sampling them from the wild, so that they are varied selectively along attributes of interest. This is enabled by recent progress in controlled and realistic image synthesis~\cite{karras2019style,karras2019analyzing}, along with methods for collecting large amounts of accurate human annotations~\cite{buhrmester2016amazon} to quantify the perceptual effect of image manipulations. 
Our synthesis approach can alter multiple attributes at a time to produce grid-like matched samples of images we call \emph{transects}. We quantify the image manipulations with detailed human annotations which we then compare with algorithm output to estimate algorithmic bias. 

We evaluate our methodology with experiments on two gender classification algorithms. We first find that our transect generation strategy creates significantly more balanced data across key attributes compared to ``in the wild'' face datasets. Next, inspired by~\cite{buolamwini2018gender}, we use this synthetic data to explore the effects of various attributes like skin color, hair length, age and perceived gender on gender classifier errors. Our findings reveal that using an experimental method can change the picture of algorithmic bias (Fig.~\ref{fig:bar-plots}), which will affect the strategy of algorithm improvement, particularly concerning groups that are often underrepresented in training and test sets. 

We view our work as a first step in developing experimental methods for algorithmic bias testing in computer vision which, we argue, are necessary to achieve trustworthy and actionable measurements. Much remains to be done, both in design and experimentation to achieve broadly-applicable and reliable techniques. In Sec.~\ref{sec-discussion-conclusions} we discuss limitations of the current method, and next steps in this research area.



\section{Related Work}
Benchmarking in computer vision has a long history~\cite{barron1994performance,bowyer1998empirical,fei2004learning} including face recognition~\cite{Krishnapriya2019,phillips2003face,phillips1998feret,phillips2018face,grother2018ongoing1,grother2018ongoing2} and face analysis~\cite{buolamwini2018gender}. Some of these studies examine biases in performance, i.e., error rates across variation of important parameters (e.g. racial background in faces). Since these studies are purely observational, they raise the question of whether the biases they measure depend on algorithmic bias, or on correlations in the test data. Our work addresses this question.

A dataset is said to be biased when combinations of features of interest are disproportionately represented or, equivalently, when such features are correlated. Computer vision datasets are often found to be biased~\cite{ponce2006dataset,torralba2011unbiased}. Human face datasets are particularly scrutinized~\cite{albiero2020analysis,drozdowski2020demographic,klare2012face,kortylewski2018empirically,kortylewski2019analyzing,merler2019diversity} because methods and models trained on these data can end up being biased along attributes that are protected by the law~\cite{kleinberg2019}. Approaches to mitigating dataset bias include collecting more thorough examples~\cite{merler2019diversity}, using image synthesis to compensate for distribution gaps~\cite{kortylewski2019analyzing}, and example resampling~\cite{li2019repair}. 

The machine learning community is active in analyzing biases of learning models, and how one may train models where bias is mitigated~\cite{alvi2018turning,corbett2018measure,das2018mitigating,hebert2017calibration,hendricks2018women,khosla2012undoing,kortylewski2019analyzing,lu2019experimental,ryu2017inclusivefacenet}, usually by ensuring that performance is equal across certain subgroups of a dataset. Here we ask a complementary question: we assume that the system to be benchmarked is {\em pre-trained} and fixed, and we ask how to reliably measure algorithmic bias in pre-trained black-box algorithms.

Studies of face analysis systems~\cite{buolamwini2018gender,klare2012face,lu2019experimental} and face recognition systems~\cite{Hanaoka2019Face-Recognitio,Krishnapriya2019} attempt to measure bias across gender and skin-color (or ethnicity). However, the evaluations are based on observational rather than interventional techniques -- and therefore any conclusions from these studies should be treated with caution. A notable exception is a recent study~\cite{muthukumar2018understanding} using the experimental method to investigate the effect of skin color in gender classification. In that study, skin color is modified artificially in photographs of real faces to measure the effects of differences in skin color, all else being equal. However, the authors observe that generalizing the experimental method to other attributes, such as hair length, is too onerous if one is to modify existing photographs.  Our goal is to develop a generally applicable and practical experimental method, where {\em any} attribute may be studied independently. 

Recent work uses generative models to explore face classification system biases. One study explores how variations in pose and lighting affect classifier performance~\cite{albiero2020analysis,kortylewski2018empirically,kortylewski2019analyzing}. A second study uses a generative model to synthesize faces along particular attribute directions~\cite{denton2019detecting}. These studies rely on the strong assumption that their generative models can modify one attribute at a time. However, this assumption relies on having unbiased training data, which is almost always not practical. In contrast, our framework uses human annotations to account for residual correlations produced by our generative model.

Finally, there is research into interpreting neural networks. One strategy is to determine regions of the input that are salient, either through analysis of gradients or perturbations of the input image~\cite{Chang2018ExplainingIC,dabkowski2017real,fong2017interpretable,goyal2019,selvaraju2017grad,simonyan2013deep,sundararajan2017axiomatic}. Network dissection approaches explore how particular neurons within a network affect the output, particularly in a semantic way~\cite{bau2019gandissect,zhou2018interpreting}. Testing with Concept Activation Vectors (TCAV)~\cite{kim2017} provides explanations at a high level using directional derivatives to reveal the ``conceptual sensitivity'' of a model's prediction of a class (e.g., Smiling) to a concept. In contrast, our approach uses a synthesis model to create carefully modified input images, and human annotations to precisely quantify them.

\section{Face Attribute Annotation in Synthetic Images}
\label{sec-attribute-annotation}
The face images used in our experiments are synthetic, and therefore there is no real person behind each image. Thus, there is no intrinsic ground truth for face attributes such as gender, hair length, and skin tone. Such attributes are instead established by human annotators. We clarify here what we mean when we talk about face attributes in the absence of a physical ground truth.

Many attributes have both intrinsic and extrinsic manifestations. For example, ``emotion'' may be studied at three levels~\cite{anderson2014framework}: an unconscious physiological state, conscious self-perception (feelings), and emotional display (e.g. facial expression)~\cite{darwin1998expression}. These quantities are {\em intrinsic} to a person's or an animal's body and are not directly accessible to a machine. By contrast, an {\em extrinsic} description, i.e., the report by an onlooker of his/her perception, are more easily accessible, and this is what the machine is trained to predict. 

Since we are using synthetic images, it should be clear that we are not attempting to access the intrinsic state of a person: there is no person, and there is no intrinsic gender, ethnicity, age or emotion. However, perception of such attributes is possible. This is the same way that onlookers instinctively classify the {\it Venus of Milo} as ``female'' and Michelangelo's {\it David} as ``male,'' despite the fact that they are idealized marble representations, rather than real people.

Thus, when we refer to the ``age'' or ``gender'' or any other attribute that is computed by a face analysis system from a picture, what we mean is {\em the algorithm's prediction of a casual observer's report of their perception of the outwards display of that attribute}. This is a bit of a mouthful, and that's why we use the abbreviated expression of ``attribute,'' ``age'' or ``gender.'' The attributes we measure from human observers are reports of subjective perceptions. However, as we find in Sec.~\ref{sec:human-annotations}, these measurements are consistent and reproducible across different observers, and so we consider statistics of such reports as objective quantities.


In our study, we discretize continuous face attributes. We have used six classes of age and skin tone, five of hair length, facial expression and gender,  etc. (see Figs.~\ref{fig:screenshots} and~\ref{fig:annotation-stats}). This choice was made to conform with the literature, e.g., the Fitzpatrick scale of skin tone~\cite{fitzpatrick1988validity}, and to accommodate the abilities of non-expert casual observers, the ``common person,'' whose perception we rely on in our experiments. We make no claim to have the perfect discretization scheme; other discretization choices may be better suited in different contexts. 

Gender deserves a special mention: {\em gender identity} is often modeled as multi-dimensional~\cite{egan2001gender}. However, here we are measuring {\em reports of gender perception} (an extrinsic variable), rather than gender identity (the intrinsic variable), and our subjects could not reliably report beyond the traditional one-dimensional M/F dimension. Therefore, following~\cite{buolamwini2018gender} we settled for one dimension, which we discretized into five steps to accommodate different levels of confidence and ambiguity. 

\section{Method}
\label{sec:causal-analysis}
Our framework consists of two components: a technique to synthesize sets of images with control over semantic attributes, and a procedure using these synthesized images, along with human annotators, to perform analysis of a recognition system. 

In Sec.~\ref{sec:transect} we present our technique for attribute-controlled image synthesis. We introduce the concept of \emph{transect}, a grid-like construct of synthesized images with a different attribute manipulated along each axis. A transect gives control over the joint distribution of synthesized attributes allowing us to generate {\em matched samples} across multiple attributes, unlike related methods that operate on only one or two attributes at a time~\cite{denton2019detecting,shen2019interpreting,singla2019explanation,xiao2018elegant}. We then collect human annotations for each transect image, to precisely quantify our modifications.

In Sec.~\ref{sec:analysis} we present analyses we can perform using the annotated transects. We report a classifier's error rate, stratified along subgroups of a sensitive attribute. We also return a covariate-adjusted estimate of the \emph{causal effect} of a binary attribute on the classifier's performance. 

\subsection{Transects: A Walk in Face Space}
\label{sec:transect}
We assume a black-box face generator $G$ that can transform a latent vector $\bz \in \mathcal{R}^D$ into an image $I = G(\bz)$, where $p(\bz)$ is a distribution we can sample from. In our study, $G$ is the generator of a pre-trained, publicly available state-of-the-art GAN (``StyleGAN2'')~\cite{karras2019style,karras2019analyzing}. GAN latent spaces typically exhibit good disentanglement of semantic image attributes. In particular, empirical studies show that each image attribute often has a direction $\bv \in \mathcal{R}^D$ that predominantly captures its variability~\cite{karras2019style,zhou2018interpreting}. We base our approach on a recent study~\cite{zhou2018interpreting} for single attribute traversals in GAN latent spaces. That method trains a linear model to predict a particular image attribute from $\bz$, and uses the model to traverse the $\bz$-space in a discriminative direction. Our method generalizes this idea to synthesize image grids, i.e., \emph{transects}, spanning arbitrarily many attributes.  

\begin{figure}[t!]
\begin{center}
\includegraphics[width=\textwidth]{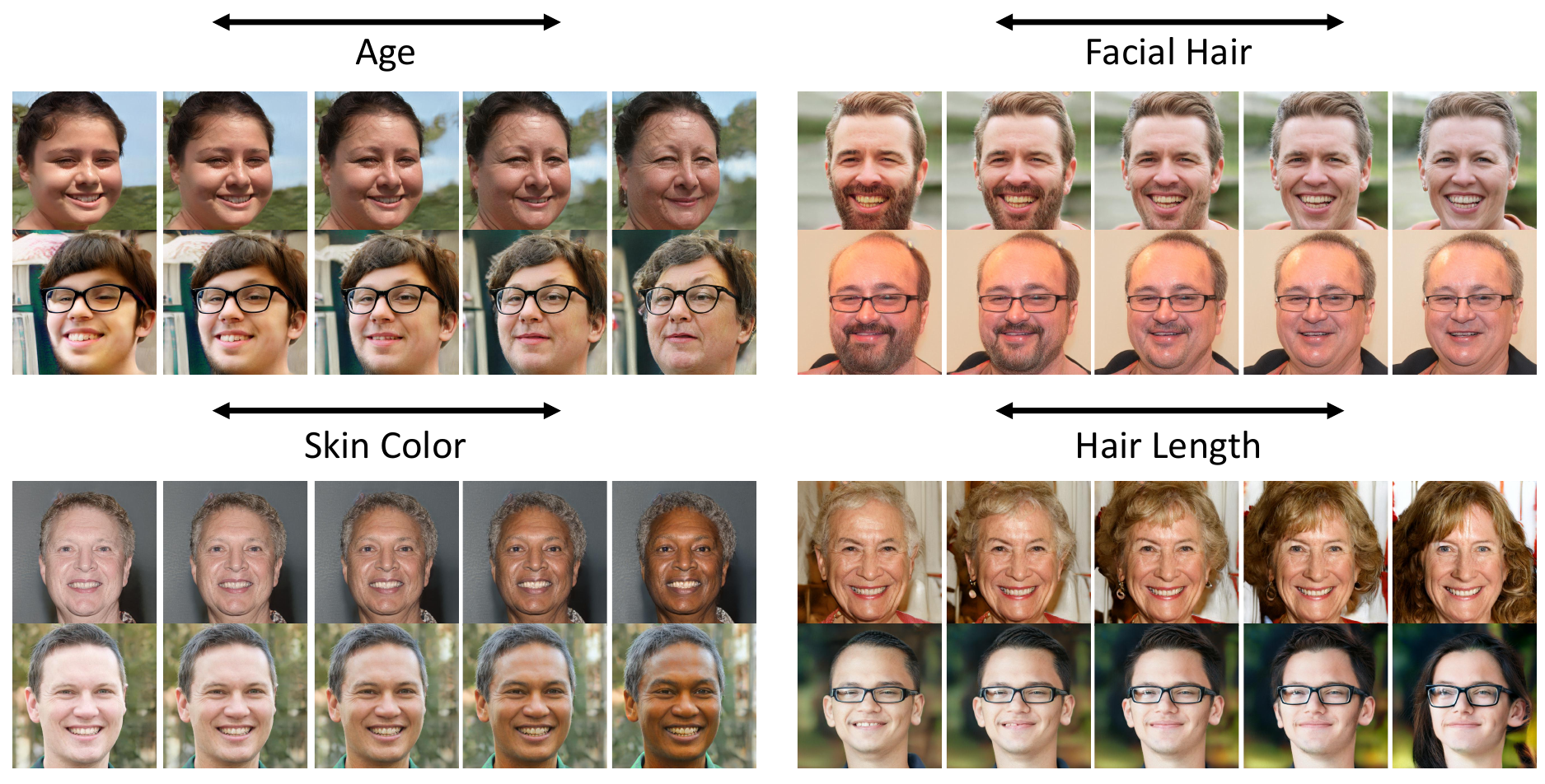}
\end{center}
\vspace{-0.3in}
\caption{\footnotesize{\bf 1D transects.} $1\times 5$ sample transects synthesized by our method for various attributes. Orthogonalization was used (see Fig.~\ref{fig:orthogonalization}). }
\label{fig:single-transect-examples}
\end{figure}

\begin{figure}[t!]
\begin{center}
\includegraphics[width=\textwidth]{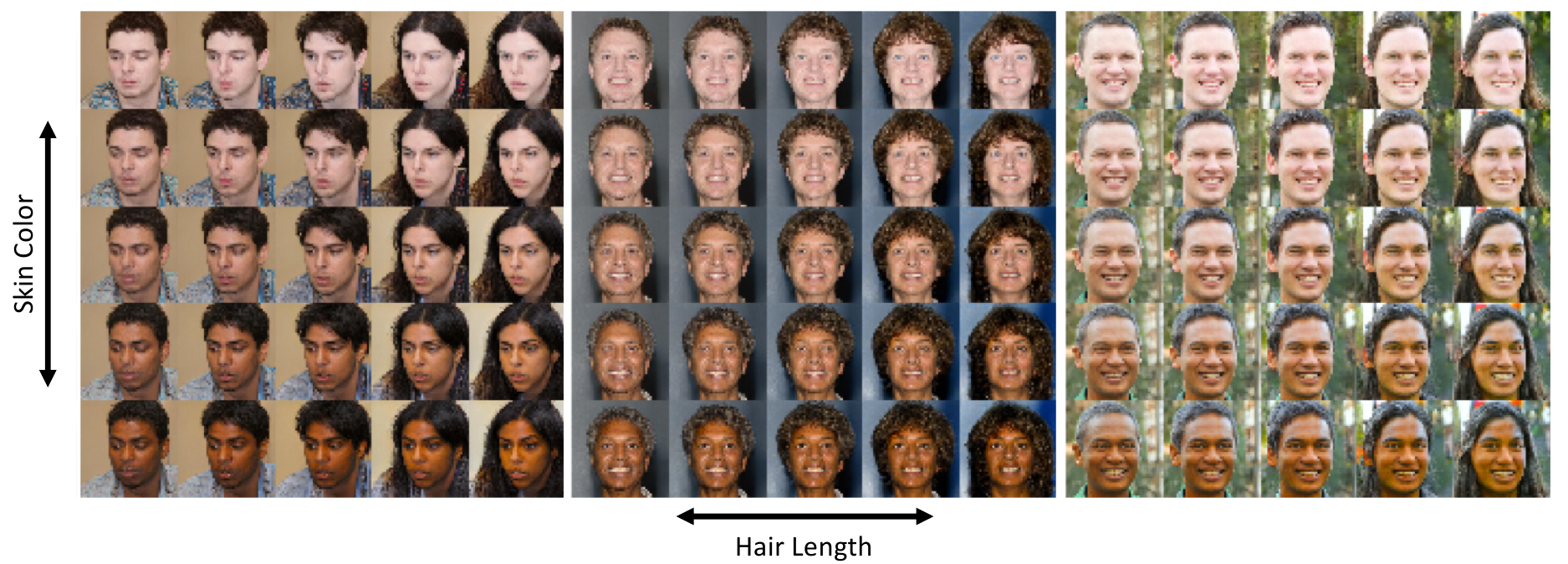}
\end{center}
\vspace{-0.3in}
\caption{\footnotesize{\bf 2D transects.} $5 \times 5$ transects varying simultaneously hair length and skin tone. Multidimensional transects allow for intersectional analysis, i.e. analysis across the joint distribution of multiple attributes. Orthogonalization was used (see Fig.~\ref{fig:orthogonalization}).}
\label{fig:multi-transect-examples}
\end{figure}

\subsubsection{Estimating Latents-to-Attributes Linear Models}
\label{sec:latent-classifier}
We first sample the latent space, measure the attributes at each location through human observers, and use these measurements to calculate principal axes of variation for attributes. More formally, let there be a list of $N_a$ image attributes of interest (age, gender, skin color, etc.). As explained below, we generate an annotated training dataset $\mathcal{D}_{\bz}=\{\bz^i, \ba^i\}_{i=1}^{N_{z}}$, where $\ba^i$ is a vector of scores, one for each attribute, for generated image $G(\bz^i)$. The score for attribute $j$, $\ba^i_j$, may be continuous in $[0, 1]$ or binary in $\{0, 1\}$. 

We produce $\mathcal{D}_{\bz}$ as follows. First, we sample a generous number of values of $\bz^i$ from $p(\bz)$. Second, we obtain labels $\ba^i$ from human annotators. A related study obtains labels by only processing the generated images through a trained classifier~\cite{zhou2018interpreting}. We generally avoid this approach because any biases of the classifier due to attribute correlations --- precisely the phenomena we are trying to avoid --- will leak into our method.

For each attribute $j$, we use $\mathcal{D}_{\bz}$ to compute a $(D-1)$-dimensional linear hyperplane $h_j = (\bn_j, b_j)$, where $\bn_j$ is the normal vector and $b_j$ is the offset. For continuous attributes like age or skin color, we train a ridge regression model~\cite{hoerl1970ridge}. For binary attributes we train a support vector machine (SVM) classifier~\cite{cortes1995support}. 

When sampling from StyleGAN2 using the native latent Gaussian distribution, we noticed a bias towards generating Caucasian-looking faces which is not surprising given the fact that it was trained on Flickr-Faces-HQ (FFHQ) -- a public dataset that is skewed towards that demographic (see Fig.~\ref{fig:violin-gender}). However, using human annotations our method is able to partially mitigate this bias by directing sampling towards the relevant portions of the latent space (see following sections), so that it could generate a diversity of attributes. Nevertheless, training face synthesis GANs with a more diverse set of faces will be an important step in making our method more easily applicable.

\subsubsection{Multi-attribute Transect Generation}
\label{sec:multi-attribute-transect}
The attribute hyperplanes may now be used to sample faces that vary along specific attributes. More formally: the hyperplane $h_j$ specifies the subspace of $\mathcal{R}^D$ with boundary or neutral values of attribute $j$, and the normal vector $\bn_j$ specifies a direction along which that attribute primarily varies. To construct a one-dimensional, length-$L$ transect for attribute $j$, we first start with a random point $\bz^i$ and project it onto $h_j$. We then query $L-1$ evenly-spaced points along $\bn_j$, within fixed distance limits on both sides of the $h_j$. Fig.~\ref{fig:single-transect-examples} presents some single transect examples (with orthogonalization, a concept introduced in the next section). We give further details on querying points in Sec.~\ref{sec:step-sizes}.

The 1D transect does not allow us to explore the joint space of several attributes, or to fix other attributes in precise ways when varying one attribute. We generalize to $K$-dimensional transects in Algorithm~\ref{alg:multi} to address this. The main extensions are: (1) we project $\bz^i$ onto the intersection of $K$ attribute hyperplanes, and (2) we move in a $K$-dimensional grid in $\bz$-space (see Fig.~\ref{fig:multi-transect-examples}). Input $\bc_k$ is a vector of decision values with respect to the hyperplane $h_k$, and $\bv_k$ is a direction vector (equivalent to $\bn_k$ here, until orthogonalization is introduced in the next section).

We are unaware of a simple closed-form solution to project $\bz^i$ onto the intersection of arbitrarily many hyperplanes. We instead take an iterative approach: we sequentially project the point onto each hyperplane, and repeat this process for some number of iterations. Repeated projections onto convex sets, the hyperplanes in our case, is guaranteed to converge to a location on the intersection of the sets~\cite{youla1987mathematical} which, in our case, is a single point. If the hyperplanes are perfectly orthogonal, this process converges in exactly one iteration; we empirically found convergence in fewer than 50 iterations. 

\begin{algorithm}[b!]
\textbf{Input}: Generator $G$, tuples $\{ (L_k, \bn_k, b_k, \bv_k, \bc_k) \}_{k=1}^K$, where $L_k$ is a transect dimension, $(\bn_k, b_k)$ is a hyperplane, $\bv_k$ is a direction vector, and $\bc_k$ are signed decision values.\\
\textbf{Output}: A $L_1 \times \cdots \times L_K$ transect $T^i$. \\ 
\texttt{\\}
 $\bz^i \sim p(\bz)$\\
 $\bz^{i,0} = $ projection of $\bz^i$ onto intersection of $\{ (\bn_k, b_k)\}_{k=1}^K$\\
 \For{$l_1=1 \cdots L_1$}{
    $\vdots$\\
    \For{$l_K=1 \cdots L_K$}{
    $T^i(l_1, \cdots , l_K) = G( \bz^{i,0} + \sum_{k=1}^K \frac{\bc_k[l_k]}{\langle \bv_k, \bn_k \rangle} \frac{\bv_k}{\|\bv_k\|})$
    }
 }
 \caption{ $K$-attribute transect generation}
 \label{alg:multi}
\end{algorithm}
\begin{algorithm}[b!]
\textbf{Input}: Vectors $\{\bn_j\}_{j=1}^{N_a}$.\\
\textbf{Output}: Vectors $\{\tilde{\bn}_j\}_{j=1}^{N_a}$, where $\tilde{\bn}_j \perp \bn_k, \forall k \neq j$ \\ 
\texttt{\\}
 $Q,R \leftarrow $ QR-factorization of matrix $[\bn_1, \bn_2, \cdots, \bn_{N_a}]$  \\
 
 \For{$i=1 \cdots N_a$}{
    $\tilde{\bn}_i = \bn_i$ \\
    \For{$j=1 \cdots N_a$}{
        \If{$i \neq j$}{
            $\tilde{\bn}_i = \tilde{\bn}_i - \frac{Q_j \cdot \langle Q_j, \tilde{\bn}_i \rangle}{\langle Q_j,Q_j \rangle}$\\
        }
    }
 }
 \caption{ Orthogonalization }
 \label{alg:ortho}
\end{algorithm}

\subsubsection{Orthogonalization of Traversal Directions}
\label{sec:orthogonalization}
The hyperplane normals $\{\bn_j\}_{j=1}^{N_a}$ are not orthogonal to one another. If we set the direction vectors equal to these normal vectors in Algorithm~\ref{alg:multi}, i.e., $\bv_j = \bn_j$, we will likely observe unwanted correlations between attributes. We reduce this effect by producing a set of modified direction vectors such that $\bv_j \perp \bn_k, \forall k \neq j$ (see Algorithm~\ref{alg:ortho}). 

\begin{figure}[h!]
    \includegraphics[width=\textwidth]{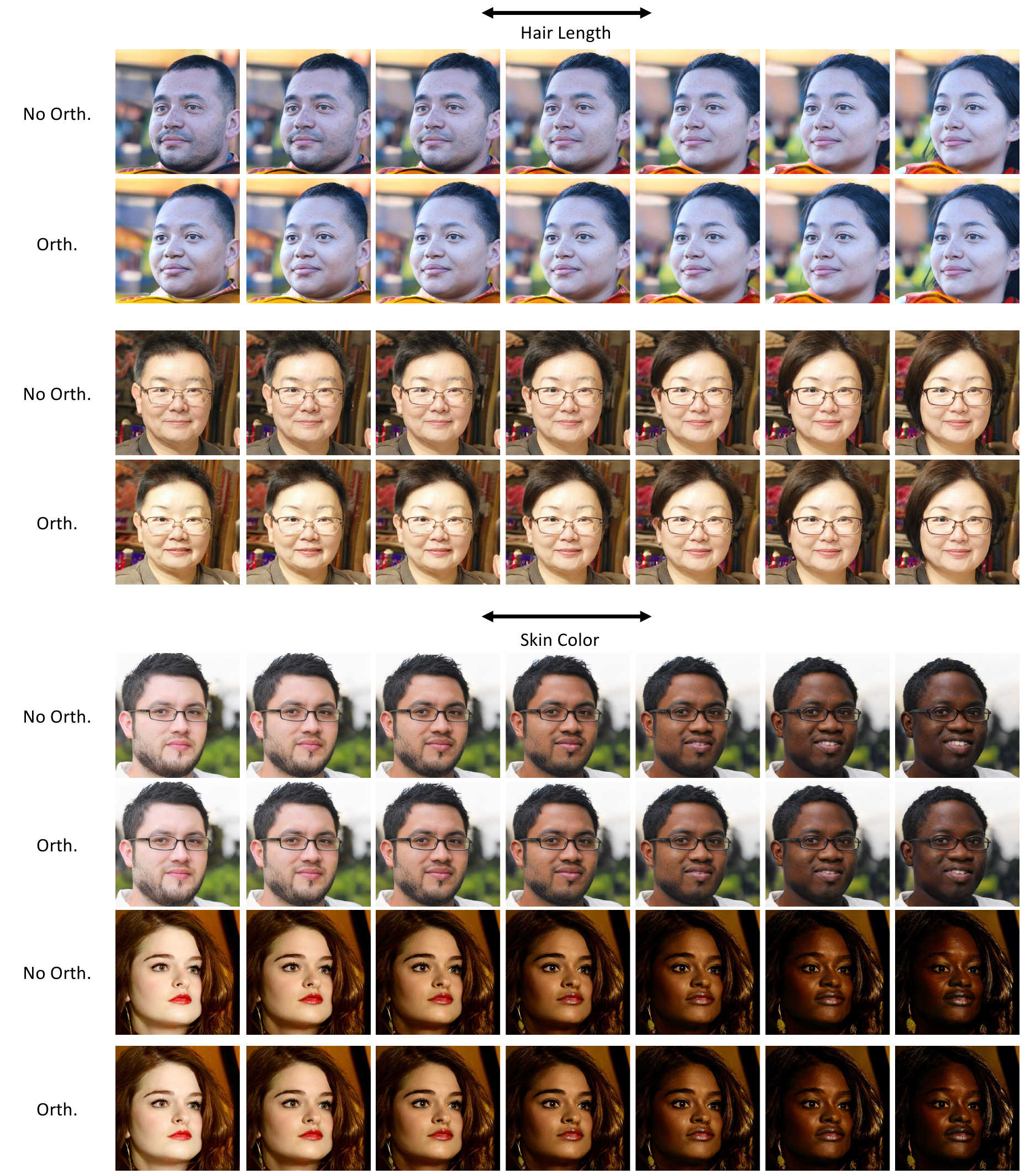}
    \caption{\footnotesize{\bf 1D transects with and without orthogonalization.} Without orthogonalization (Sec.~\ref{sec:multi-attribute-transect}), decreasing hair length results in more masculine-looking faces. This phenomenon is not as apparent after orthogonalization (Sec.~\ref{sec:orthogonalization}). We see only slight orthogonalization differences in the skin color transects.}
\label{fig:orthogonalization}
\end{figure}

Fig.~\ref{fig:orthogonalization} illustrates the effects of orthogonalization for hair length and skin color. Without orthogonalization, the hair length transects exhibit unwanted changes in gender, with shorter hair also causing faces to appear more masculine. With orthogonalization, these unwanted changes are removed. In contrast, we see no clear difference in skin color transects with and without orthogonalization, indicating that the skin color hyperplane was already near-orthogonal to the other attribute hyperplanes. 

\subsubsection{Setting Step Sizes and Transect Dimensions}
\label{sec:step-sizes}
If human annotation cost were negligible, we could simply query many grid locations with large transect dimensions $L$ to capture subtle appearance changes over the dynamic ranges of the attributes. But given constrained resources, we set $L$ to small values. For example, $L=5$ for the 1D transects in Fig.~\ref{fig:single-transect-examples} and 2D transects in Fig.~\ref{fig:multi-transect-examples}, and $L=2$ for the 3D transects in Fig.~\ref{fig:octets}. For each attribute $j$, we manually set min/max signed decision values with respect to $h_j$, and linearly interpolate $L_j$ points between these extremes to obtain $\bc_j$. We set per-attribute min/max values so that transects depict a full dynamic range for most random samples.

\subsection{Analyses Using Transects}
\label{sec:analysis}
We assume a target attribute of interest, e.g., gender, and a target attribute classifier $C$. We will use transect images to perform bias analysis on $C$. Though an ideal transect will modify only selected attributes at a time, in practice, unintended attributes may also be accidentally modified. In addition, the degree to which an attribute is altered varies across transects. To measure and control for these factors we annotate each image of each transect, resulting in a second dataset \mbox{$\mathcal{D}_{transect}=\{I^i, \ba^i\}_{i=1}^{N_{images}}$} of images and human annotations.

We denote the ground truth gender of image $I^i$ (as reported by humans) by $y^i$, and $C$'s prediction by $\hat{y}^i$. For ease of analysis, we discretize the remaining attributes into bins, and assign an independent binary variable to each bin~\cite{gelman2006data}. For instance, we may represent the `skin color' attribute with six binary variables, corresponding to the six levels shown in Fig.~\ref{fig:screenshots} (top right). For a given image, only one of these six variables would be set to 1 -- often called a `one-hot encoding.' We denote the vector of concatenated binary covariates for image $i$ by $\bx^i$, and the classification error by $e^i = \ell (\hat{y}^i, y^i)$, where $\ell(\cdot, \cdot)$ is an error function.

Our first analysis strategy is to simply compare $C$'s error rate across different subgroups in the population. Let $E_j^s$ denote the average error of $C$ over test samples for which covariate $j$ is equal to $s \in \{0, 1\}$: 

\small
\begin{align}
E_j^s = \frac{\sum_i e^i \mathbbm{1} (\bx_j^i = s)}{ \sum_i \mathbbm{1} (\bx_j^i = s)}.
\end{align}
\normalsize
\noindent If the data is generated from a perfectly randomized or controlled study, the quantity $E_j^1 - E_j^0$ is a good estimate of the ``average treatment effect'' (ATE)~\cite{angrist1995identification,heckman2001instrumental,oreopoulos2006estimating,rubin2006matched} of covariate $j$ on $e$, or the average change in $e$ over all examples when covariate $j$ is flipped from $0$ to $1$, with other covariates fixed. For example, the ATE of the ``dark skin'' covariate captures the average change in $C$'s error when each person's skin tone is changed from non-dark to dark. Exactly computing the ATE from an observational dataset is not possible, because we do not observe the counterfactual case(s) for each data point, e.g., the same person with both light and dark skin tones.

Though our transects come closer to achieving an ideal controlled study than do observational datasets ``from the wild'' (see Sec.~\ref{sec:real-comparison} for empirical validation), there may still be some confounding between covariates in practice (see Fig.~\ref{fig:hair-and-beard} for an example). Since any observable confounder may be annotated in $\mathcal{D}_{transect}$, we can employ covariate-adjusted ATE estimators~\cite{pocock2002subgroup, robinson1991some, willan2004regression}. One simple covariate adjustment approach is to train a linear regression model predicting $e^i$ from $\bx^i$:
\small
\begin{align}
    e^i =  \epsilon^i + \beta_0 + \sum_{j}\beta_j \bx^i_j,
\end{align}
\normalsize
where $\beta$'s are parameters, and $\epsilon^i$ is a per-example noise term. $\beta_j$ captures the ATE, the average change in $e$ given one unit change in $\bx_j$ holding all other variables constant, provided: (1) a linear model is a reasonable fit for the relationship between the dependent and independent variables, (2) all relevant attributes are included in the model (i.e., no hidden confounders), and (3) no attributes that are influenced by $\bx_j$ are included in the model, otherwise these other factors can ``explain away'' the impact of $\bx_j$. 

An experimenter can never be completely sure that (s)he has satisfied these conditions but (s)he can strive to do so through careful consideration. Discretizing and binarizing attributes helps with (1), though we still found some non-linear influences between covariates in our experiments (see Sec.~\ref{sec:joint}). As an example of (2), we found that earrings may be an important attribute that we did not account for in our analysis (see Fig.~\ref{fig:earrings}). 

Finally, when the outcome lies in a fixed range, as is the case in our experiments with $e^i \in [0, 1]$, we use logistic instead of linear regression. $\beta_j$ then represents the expected change in the \emph{log odds} of $e$ for a unit change in $\bx_j$. We use such a logistic regression analysis in our experiments (see Sec.~\ref{sec:regression}). 

\subsection{Human Annotation}
\label{sec:human-annotations}
We collect human annotations on the synthetic faces to construct $\mathcal{D}_{\bz}$ and $\mathcal{D}_{transect}$. The annotators were recruited on Amazon Mechanical Turk~\cite{buhrmester2016amazon} through the AWS SageMaker Ground Truth service~\cite{groundTruth}.  Annotators evaluated each image for seven attributes: gender, facial hair, skin color, age, makeup, smiling, hair length and image fakeness. Each attribute was evaluated on a discrete scale. Each annotator evaluated each image for one attribute at a time. For each image, we collected 5 annotations per attribute for a total of 40 annotations per image. 

\begin{figure}[ht!]
\centering
\includegraphics[width=\textwidth]{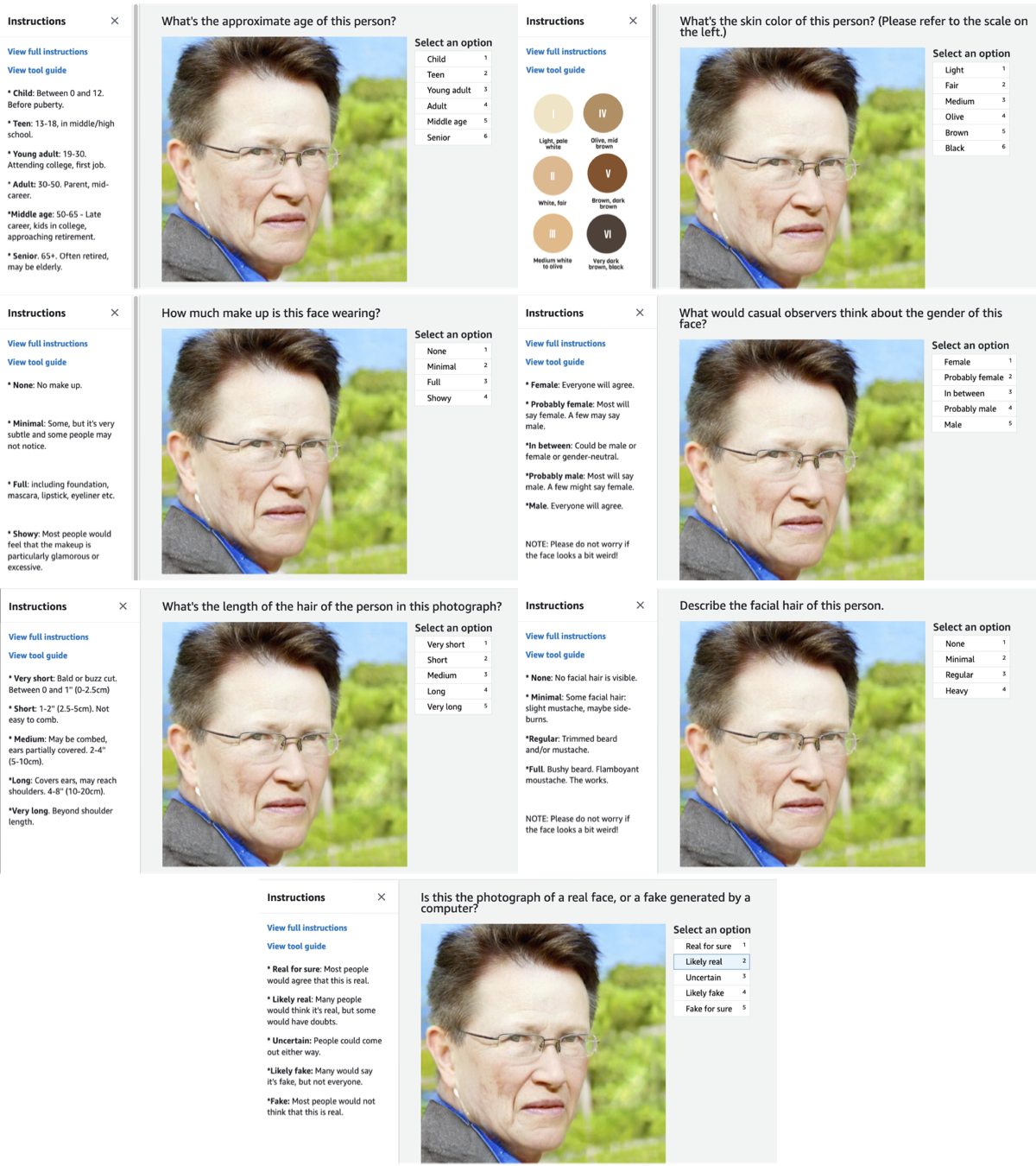}
\caption{\footnotesize {\bf Screenshots of the graphical user interface} for seven annotations we collected from Amazon Mechanical Turk annotators using the SageMaker Ground Truth service~\cite{groundTruth}.}
\label{fig:screenshots}
\end{figure}

We discretized each attribute using three to six levels. For example, we use the Fitzpatrick six-point scale for skin color~\cite{fitzpatrick1988validity}, and split age into six groups ranging from children to senior citizens. For complete details about subgroups for each attribute, along with samples of our Mechanical Turk survey layouts please see Fig.~\ref{fig:screenshots}.

\begin{figure}[t!]
    \begin{subfigure}{\textwidth}
    \centering
    \includegraphics[width=\textwidth]{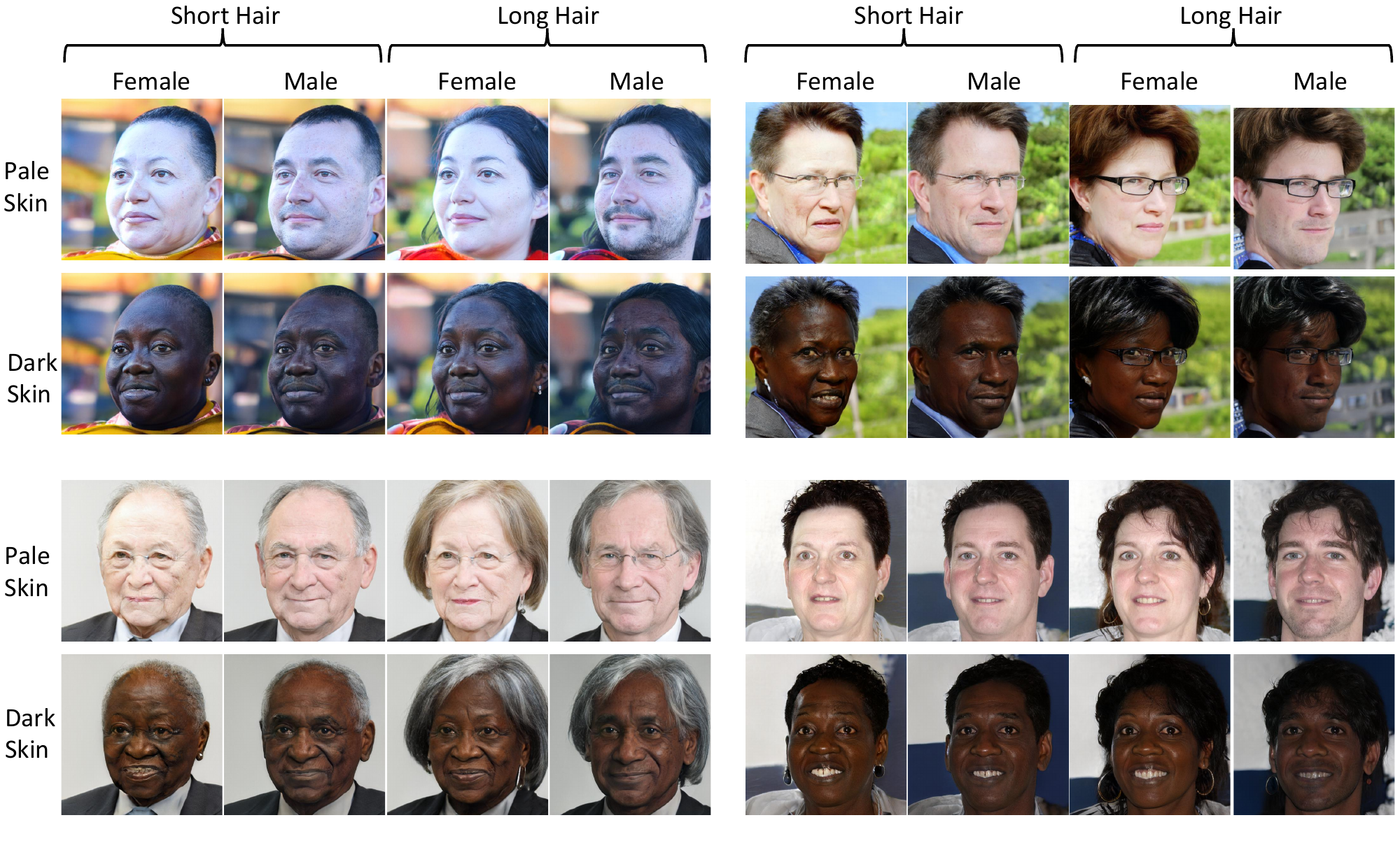}
    \vspace{-0.8cm}
    \caption{\footnotesize \bf{Examples of transects used in our experiments.}}
    \end{subfigure}
    \begin{subfigure}{\textwidth}
    \centering
    \vspace{0.5cm}
    \includegraphics[width=\textwidth]{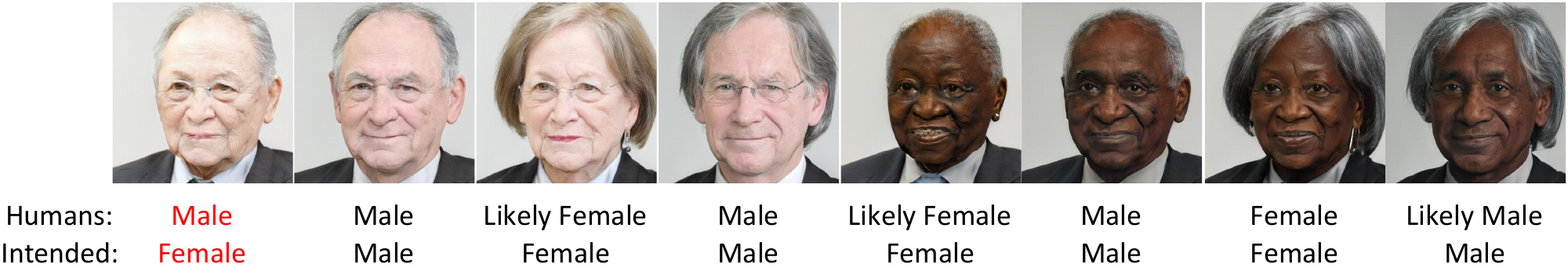}
    \caption{ \footnotesize \bf{Human perception of the generators' manipulations.}}
    \end{subfigure}
\caption{\footnotesize {\bf Sample of 3-attribute transects used in our experiments.} We created 1,000 $2 \times 2 \times 2$ transects spanning skin color, hair length and gender -- four examples are shown in (a). We set step sizes in such a way that we obtained pale-to-dark skin tones, short-to-medium hair lengths 
and M/F gender (see Sec.~\ref{sec:step-sizes}). Besides the intentionally modified attributes, other face attributes are held constant. For each image in each transect we collected human annotations to measure the perceived attributes. In (b) we show human-annotated gender values of the bottom-left transect in (a) side-by side with the generator's intended values. Humans label the first face as a male, though the generator intended to produce a female. In all our experiments we used human perception, rather than intended generator attributes, as the ground truth.}
\label{fig:octets}
\end{figure}

The number of annotations that are needed by our method is rather formidable. However, we found that this is not an obstacle in practice. In our experiments, $\mathcal{D}_{\bz}$ consists of 5,000 images, and $\mathcal{D}_{transect}$ consists of 1,000 8-image transects (see examples in Fig~\ref{fig:octets}). The total number of annotations was thus 13,000 (images) x 8 (attributes) x 5 (annotations per image and per attribute) = 0.52M  annotations. Amazon Mechanical Turk delivered on average 10-20 annotations per second, thus annotations took about 10 hours to complete over two separate sessions. Annotators were paid 1.2c per annotation, earning 10-15 US\$ per hour.

\begin{figure}[ht!]
    \centering
    \includegraphics[width=\textwidth]{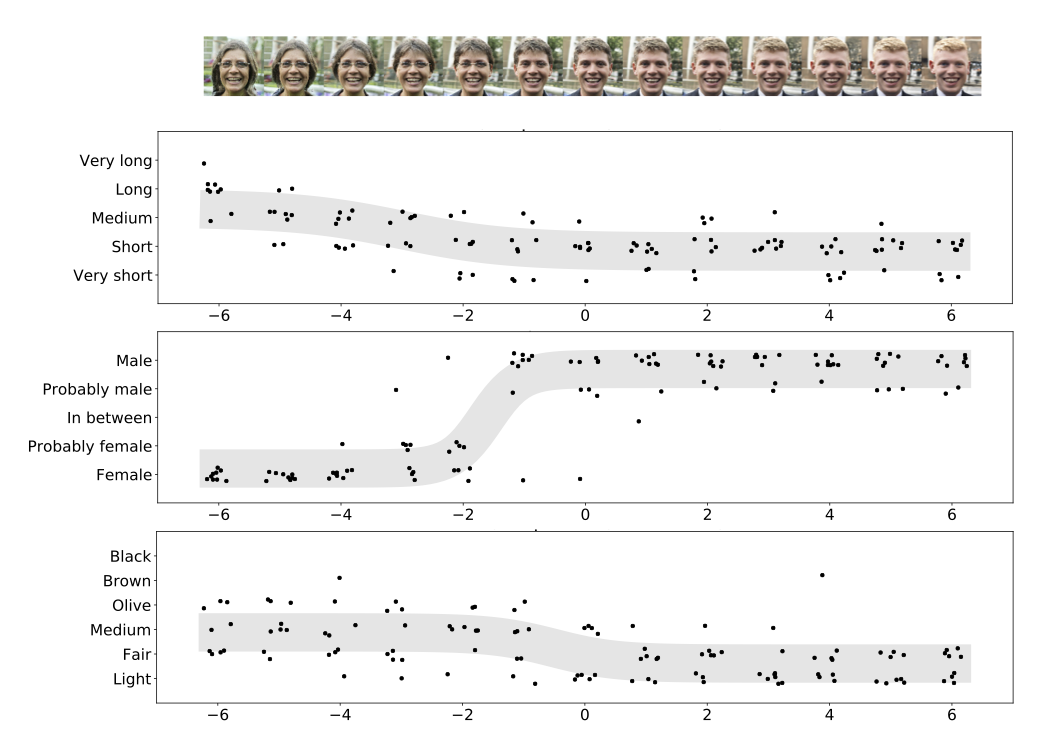}\\
    \caption{\footnotesize{\bf Annotation consistency.} Hair length  (top), gender (middle) and skin tone (bottom) annotations on a 13-image 1D transect. This transect was annotated in a pilot experiment to fine tune our GUIs and to evaluate the consistency of the annotators, and not used in our main experiment. Here nine annotations were obtained for each attribute and for each image. The annotations are shown as dots below each image. The $x$ axis increments one unit from one image to the next. A small amount of noise was added in $x$ and $y$ in order to visualize the individual annotations. The thick gray curves show the fit of a logistic function to the data. Annotations typically fall within one or two neighboring attribute levels. There are very few outliers. For a quantitative overall analysis see Fig.\ref{fig:annotation-stats}.}
    \label{fig:annotated-long-transects}
\end{figure}

\begin{figure}[t!]
\includegraphics[width=\textwidth]{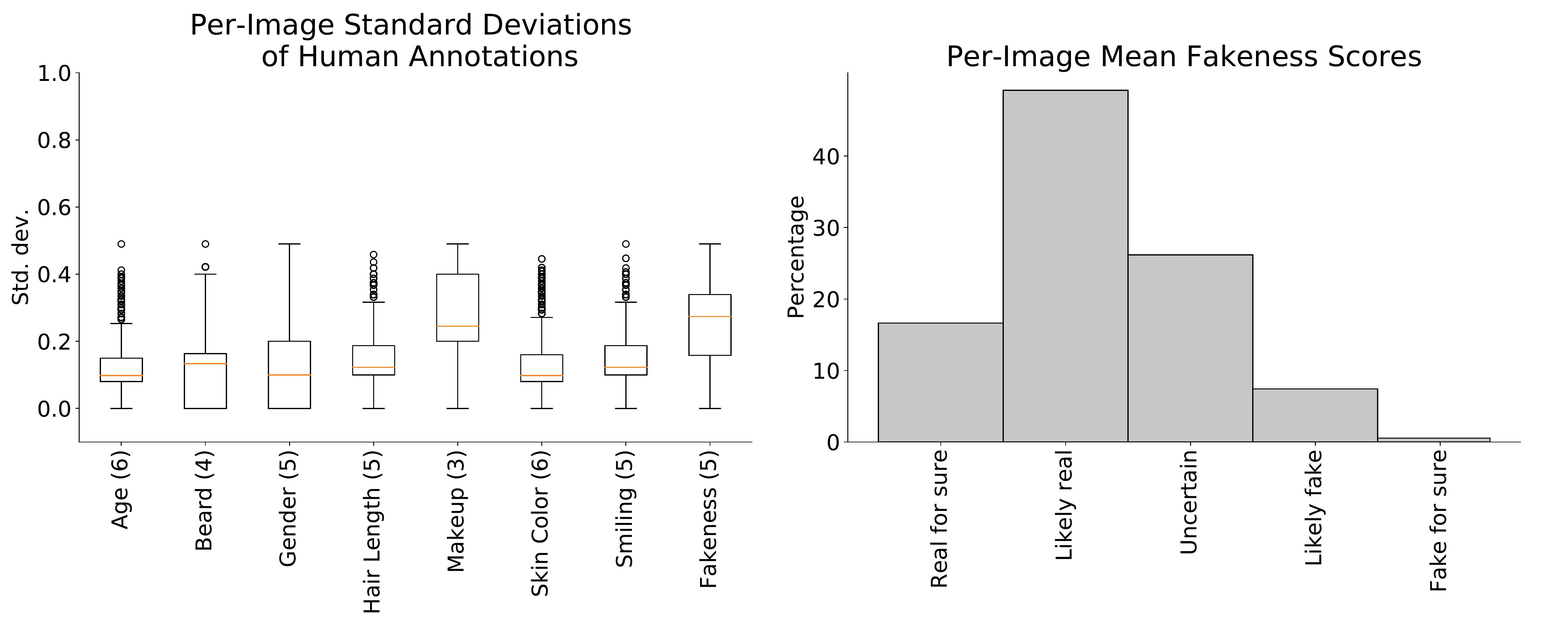}\\
\vspace{-0.3in}
\includegraphics[width=\textwidth]{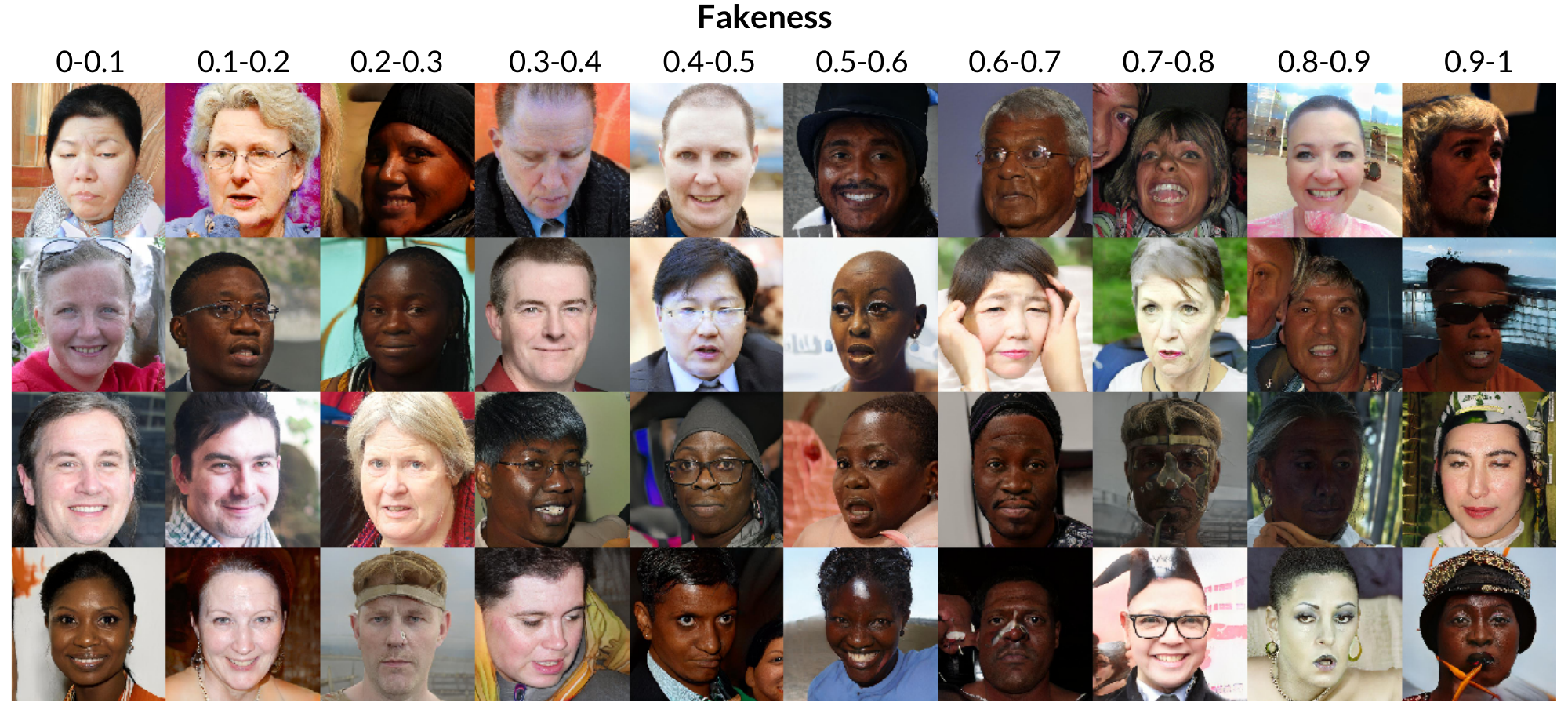}\\
\caption{\footnotesize{\bf Annotation quality and image realism.} (Top left) Distributions of per-image standard deviations of human annotations for each of the attributes we considered (one unit = dynamic range of the attribute). Five annotators were asked to provide a rating for each attribute of each image. The number of rating options per attribute is indicated in brackets next to the attribute's name. The median standard deviations (red lines) are comparable to the quantization step, indicating good annotator agreement. (Top right) We asked our annotators to rate the realism of the images. The distribution of such scores is shown. Fewer than 10\% of the ratings indicated fake or likely fake, suggesting that the synthetic images we randomly sampled are fairly realistic. (Bottom) we show examples of synthesized faces organized by mean human fakeness scores. Images with high fakeness scores were removed from the experiments (see Sec.\ref{sec:pruning}).}
\label{fig:annotation-stats}
\end{figure}

One may be concerned that annotators may not be able to give meaningful attribute annotations on synthetic images. Therefore  we explored the level of agreement of annotator responses, both in a number of pilot experiments, and in the annotations we collected for the main experiment. Fig.~\ref{fig:annotated-long-transects} shows the raw annotations for one 1D transect and three attributes. One may see that there are very few outlier annotations, and that in most cases annotations fall in one or two neighboring attribute levels. 
Fig.~\ref{fig:annotation-stats} (top left) shows a distribution of per-image annotation standard deviations, split by attribute. One unit corresponds to the dynamic range of each attribute. For most attributes, the median annotator standard deviation is near 0.1, i.e. less than the separation between attribute levels. These observations indicate good agreement between annotators and suggest that annotations are meaningful and reproducible. 

Fig.~\ref{fig:annotation-stats} (top right) presents the distribution of mean annotator fakeness scores for the synthesized images. Only a small portion of images are deemed ``Likely fake'' or ``Fake for sure.'' Realism of images is particularly important in our analysis, since image artifacts can unknowingly affect the decisions of gender classifiers. In our experiments, we remove images with a fakeness score above a certain threshold (see Sec.~\ref{sec:pruning}). Fig.~\ref{fig:annotation-stats} (bottom) shows example synthesized images organized by mean human fakeness score.

\section{Experiments}
\label{sec-experiments}

In order to test our method on a practical application, we experiment with benchmarking bias of gender classifiers. The Pilot Parliaments Benchmark (PPB)~\cite{buolamwini2018gender}, a dataset of faces of parliament members of various nations, was the first wild-collected test dataset to balance gender and skin color with the goal of fostering the study of gender classfication bias across both attributes. The authors of that study found a much larger error rate on dark-skinned females, as compared to other groups and conjectured that this is due to bias in the algorithms, i.e., that the performance of the algorithm changes when gender and skin color are changed, all else being equal. Our method allows us to test this hypothesis.

\subsection{Gender Classifiers}
We trained two research-grade gender classifier models, each using the ResNet-50 architecture~\cite{he2016deep}. The first was trained on the CelebA dataset~\cite{liu2015faceattributes}, and the second on the FairFace dataset~\cite{karkkainen2019fairface}. CelebA is the most popular public face dataset due to its size and rich attribute annotations, but is known to have severe imbalances~\cite{denton2019detecting}. The FairFace dataset was introduced to mitigate some of these biases.

We trained our classifiers for 20 epochs with the binary cross-entropy loss. We set the learning rate at $1e^{-4}$ for the first 10 epochs, and $1e^{-5}$ for the final 10 epochs. To avoid a baseline bias of predicting one gender over another, we enforced the likelihood of sampling male and female faces during training to be equal.

We decided not to test commercial system for two reasons. First, reproducibility --- the models we test may be re-implemented and re-trained by other researchers at any time, while commercial systems are black boxes which may change unpredictably over time. Second, our ResNet-50 models show biases comparable to those observed in the original study by~\cite{buolamwini2018gender} (see Fig.~\ref{fig:teaser}.

\subsection{Transect Data}
\label{sec:transect-data}
To produce the synthetic images for our transects, we used the generator from the StyleGAN2 architecture trained on Flickr-Faces-HQ (FFHQ)~\cite{karras2019style,karras2019analyzing}. This generator has both a multivariate Normal input noise space, $\mathcal{N}(\mathbf{0}, \mathbf{I})$, as well as an intermediate ``style space.'' To train the latent space linear models (see Sec.~\ref{sec:latent-classifier}), we sampled $5000$ vectors from the noise distribution, and labeled the generated images with human annotators (see Sec.~\ref{sec:human-annotations}). However, we use the \emph{style space} as the latent space in our method, because we found it better suited for disentangling semantic attributes. We trained linear regression models to predict age, gender, skin color and hair length attributes from style vectors. For the remaining attributes --- facial hair, makeup and smiling --- we found that binarizing the ranges and training a linear SVM classifier works best.

We generated 3D transects across subgroups of skin color, hair length, and gender following the procedure described in Sec.~\ref{sec:multi-attribute-transect}. We use a transect size of $2 \times 2 \times 2$, with grid decision values (specified by input vector $\bc$ in Algorithm~\ref{alg:multi}) spaced to generate a pale-to-dark transition along the skin color axis, short-to-medium length along the hair length axis, and male-to-female along the gender axis. We set the decision values by trial-and-error, and made them equal for all transects: $(-1.5, 1.7)$ for skin color, $(-0.5, 0)$ for hair length, and $(-1.75, 1.75)$ for gender. We generated 1000 such transects, resulting in $8000$ total images. Fig.~\ref{fig:octets} presents four example transects. The general characteristics of the faces --- besides the intentionally modified attributes --- are held constant. 

\begin{figure}[t!]
\centering
\includegraphics[width=0.90\textwidth]{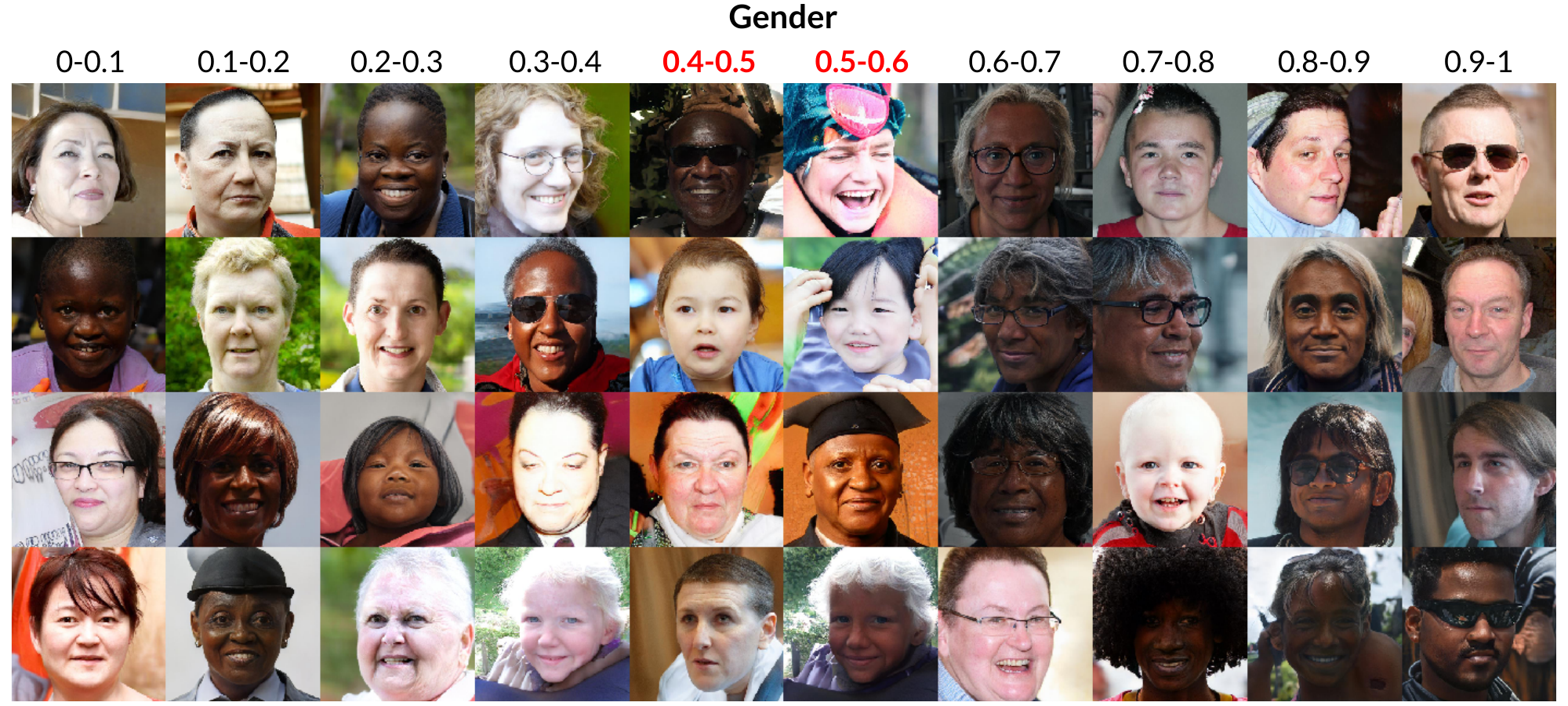}\\
\includegraphics[width=0.90\textwidth]{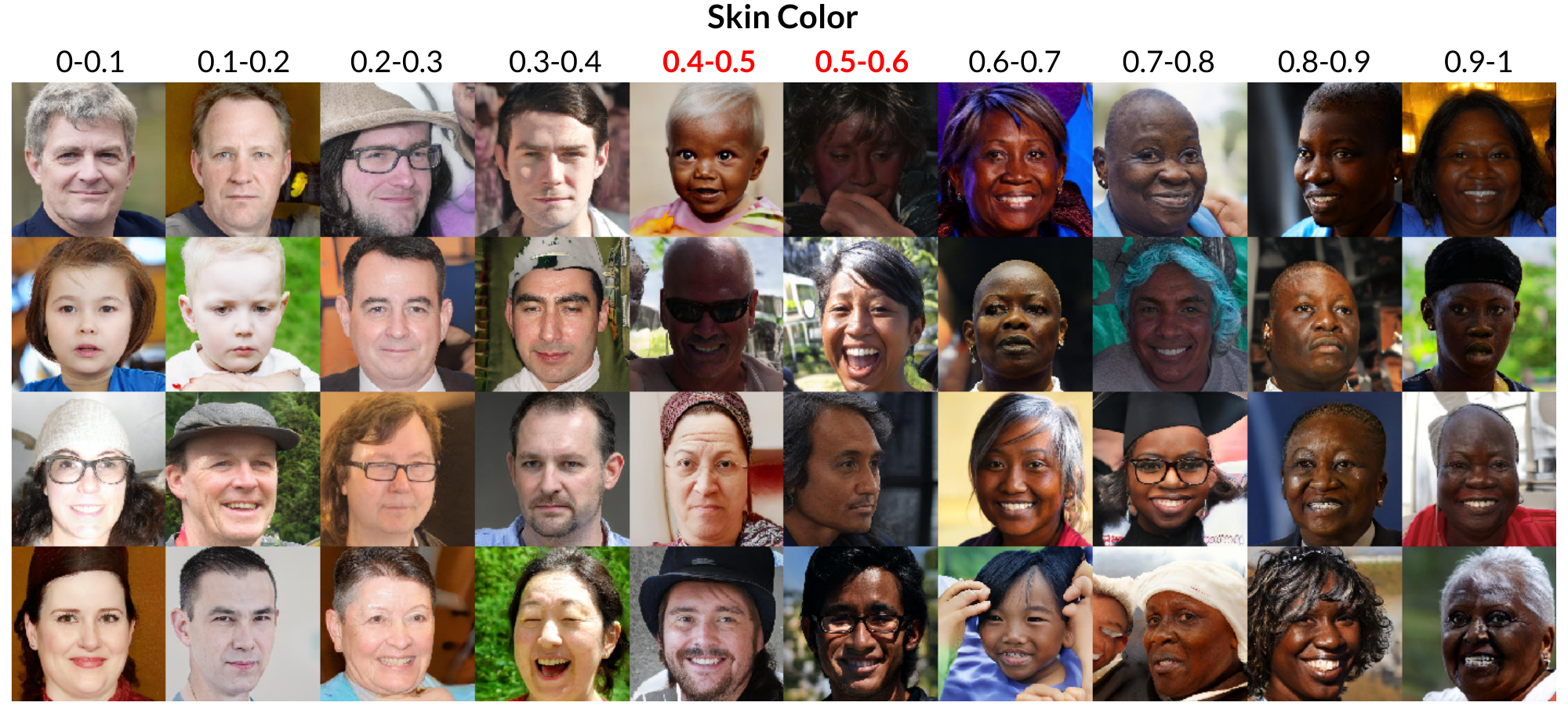}\\
\includegraphics[width=0.90\textwidth]{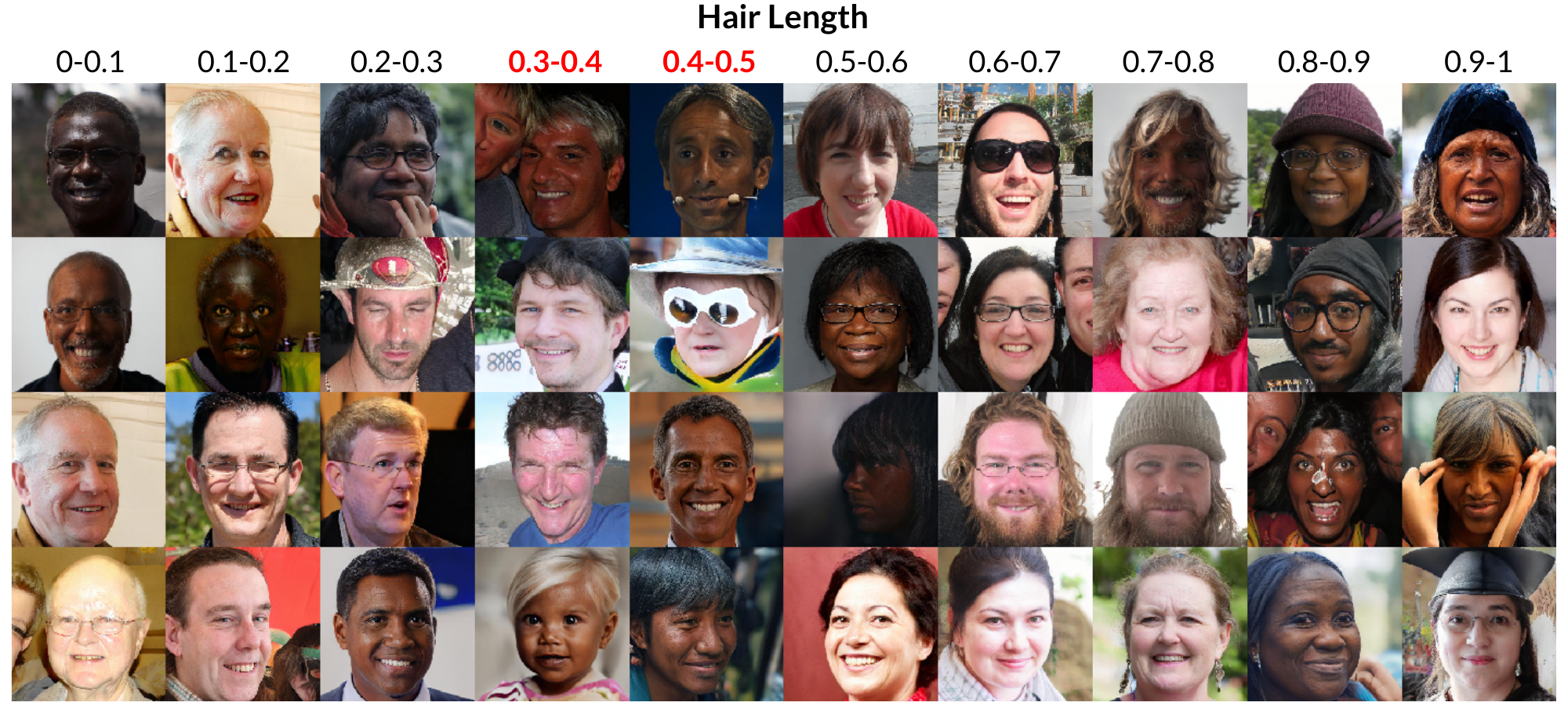}\\
\vspace{-0.1in}
\caption{\footnotesize {\bf Samples of synthesized faces, organized by mean human annotation scores.} In our analysis, we omitted faces from ranges indicated in red to focus on clearly perceived females/males, light/dark skin tones, and short/long hair lengths.}
\vspace{-0.2in}
\label{fig:omission}
\end{figure}

\subsubsection{Dataset Pruning}
\label{sec:pruning}
Not all synthesized images are ideal for our analysis. Some elicit ambiguous human responses (Fig.~\ref{fig:annotation-stats} top left) or are unrealistic (Fig.~\ref{fig:annotation-stats} top right). Furthermore, others may not belong clearly to one of our two intended categories for the gender, hair length, and skin color attributes. We addressed these points by first removing any image with a mean fakeness score greater than or equal to ``Likely fake'' ($0.75$ in the normalized range of $[0,1]$). We also removed faces with attribute values in the normalized subranges of $[0.4, 0.6]$ for skin color and gender, and $[0.3, 0.5]$ for hair length (see Fig.~\ref{fig:omission} for examples). After these pruning steps, we were left with 5713 images. 

\begin{figure}[t!]
\includegraphics[width=\textwidth]{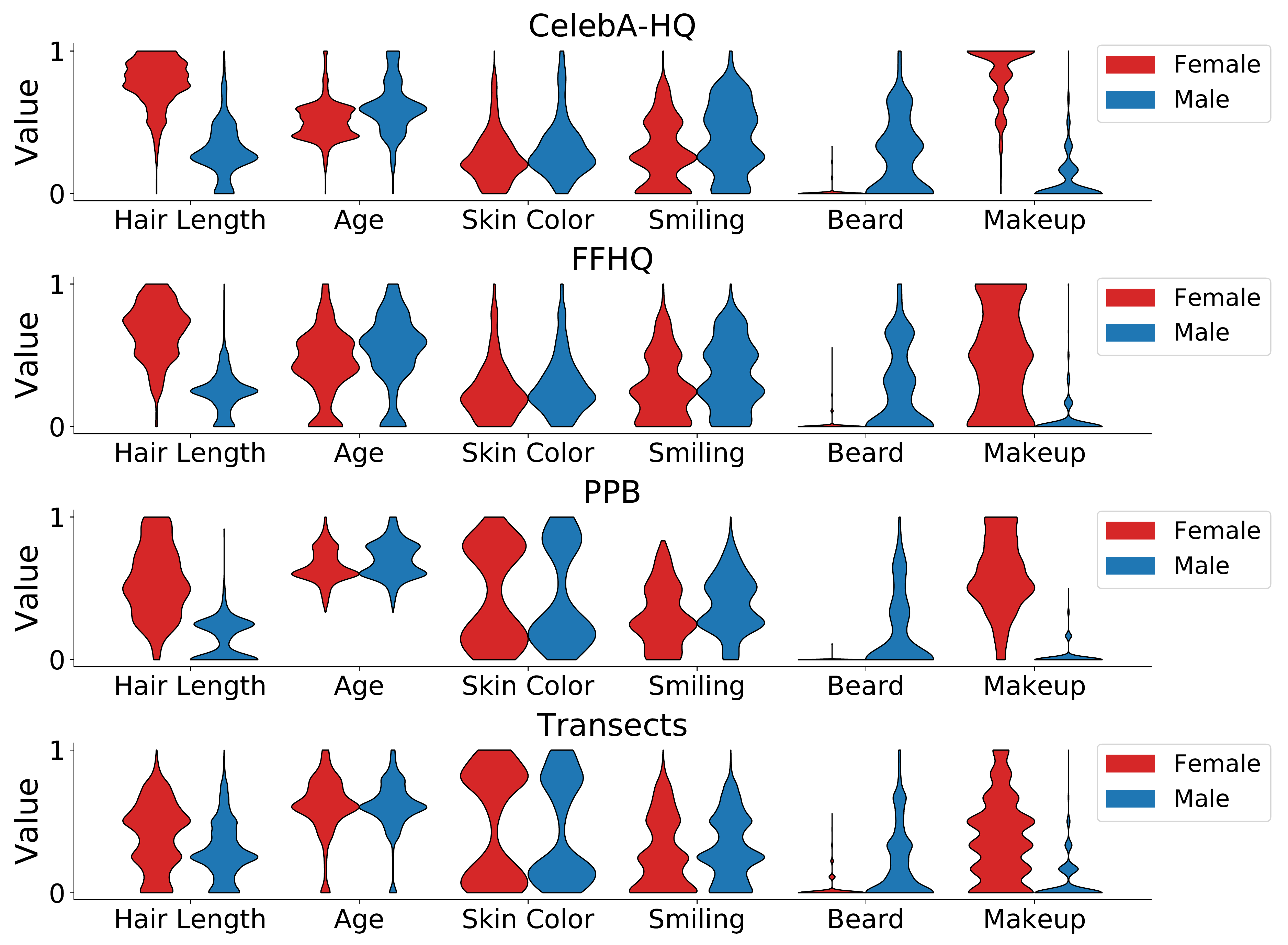}
\caption{\footnotesize{\bf Attribute distributions by dataset and gender groups.} ``Violin'' plot widths are proportional to frequency counts, and each violin is scaled so that its maximum count spans the full width. Wild-collected datasets have greater attribute imbalances across gender than synthetic transects, e.g. longer hair and younger ages for women. We designed our transects to mirror PPB skin color distribution and age distributions, while mitigating hair length imbalance. Hair length vs. skin color distributions are further explored in Fig.\ref{fig:violin-color}. }
\label{fig:violin-gender}
\vspace{0.2in}
\includegraphics[width=\textwidth]{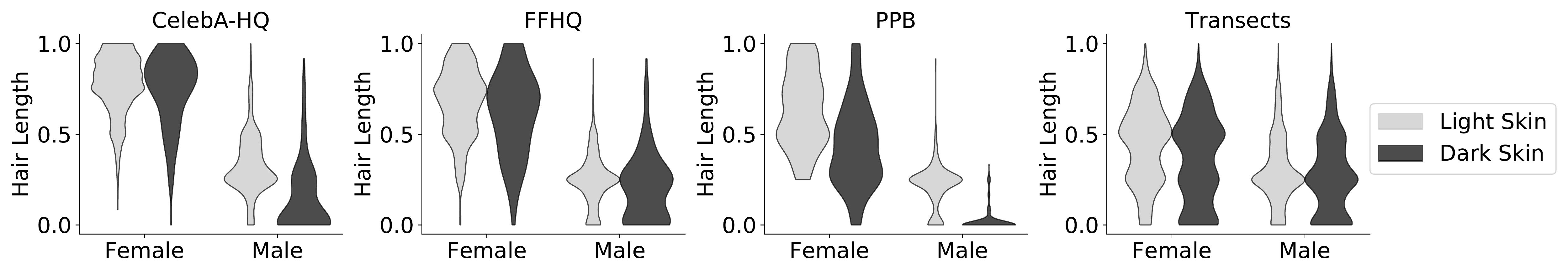}
\caption{\footnotesize{\bf Hair length distributions by gender and skin color groups.} In the wild-collected datasets hair length is correlated with skin color, when gender is held constant. Synthetic transects may be designed to minimize this correlation.}
\label{fig:violin-color}
\end{figure}

\subsection{Comparison of Transects to Real Face Datasets}
\label{sec:real-comparison}
Fig.~\ref{fig:violin-gender} analyzes attribute distributions for the CelebA-HQ, FFHQ and PPB datasets, along with our transects, stratified by gender. The wild-collected datasets contain significant imbalances across gender, particularly with hair length. They also have biases in age, with a larger percentage of males being older than females. An interesting correlation is that males are also more likely to smile in these data. In contrast, our transects exhibit more balance across gender. They depict more males with medium-to-long hair, and fewer females with very long hair. Our transects also have a bimodal skin color distribution, and an older population by design, since we are interested in mimicking those population characteristics of PPB. All datasets are imbalanced along the ``Beard'' and ``Makeup'' attributes --- this is reasonable since we expect these to have strong correlations with gender.

In an ideal matched study, sets of images stratified by a sensitive attribute will exhibit the same distribution over remaining attributes. Put simply, no other attribute should be strongly correlated with the attribute being manipulated. Fig.~\ref{fig:violin-color} stratifies by skin color. We see correlations of hair length distributions and skin colors in all the wild-collected data, while the synthetic transects exhibit much better balance. 

\begin{figure}[ht!]
\centering
\includegraphics[width=0.90\textwidth]{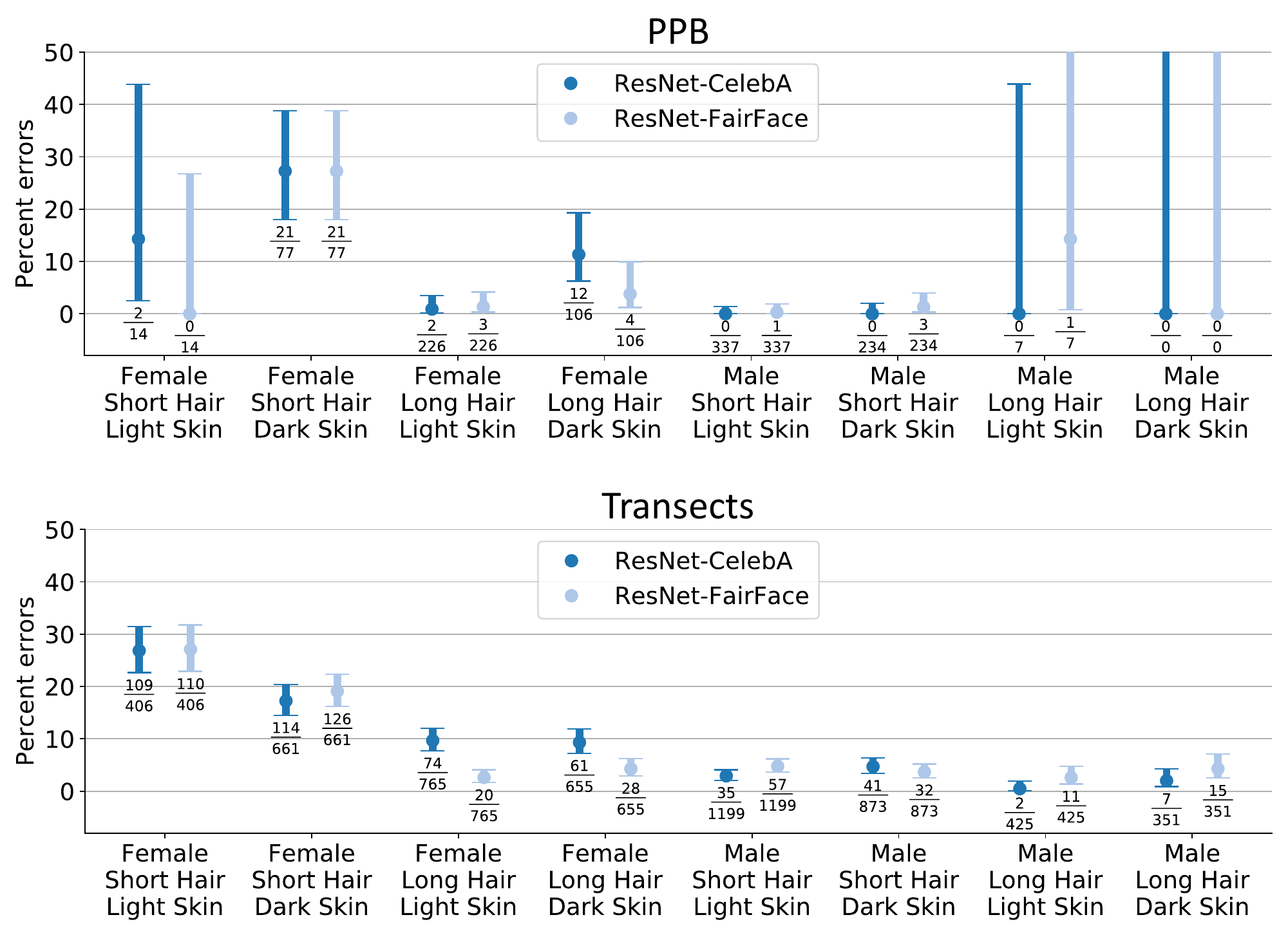}
\vspace{-0.1in}
\caption{\footnotesize {\bf Algorithmic errors, disaggregated by intersectional groups for wild-collected (PPB, top) and synthetic (transects, bottom).} Wilson score 95\% confidence intervals~\cite{wilson1927probable} are indicated by vertical bars, and the misclassification count and total number of samples are written below each bar. PPB has few samples for several groups, such as short-haired, light-skinned females and long-haired males (see Fig.~\ref{fig:violin-color}). Synthetic transects provide numerous test samples for all groups. The role of the different attributes in causing the errors is studied in Fig.~\ref{fig:regression-results}.}
\label{fig:bar-plots}
\includegraphics[width=0.99\textwidth]{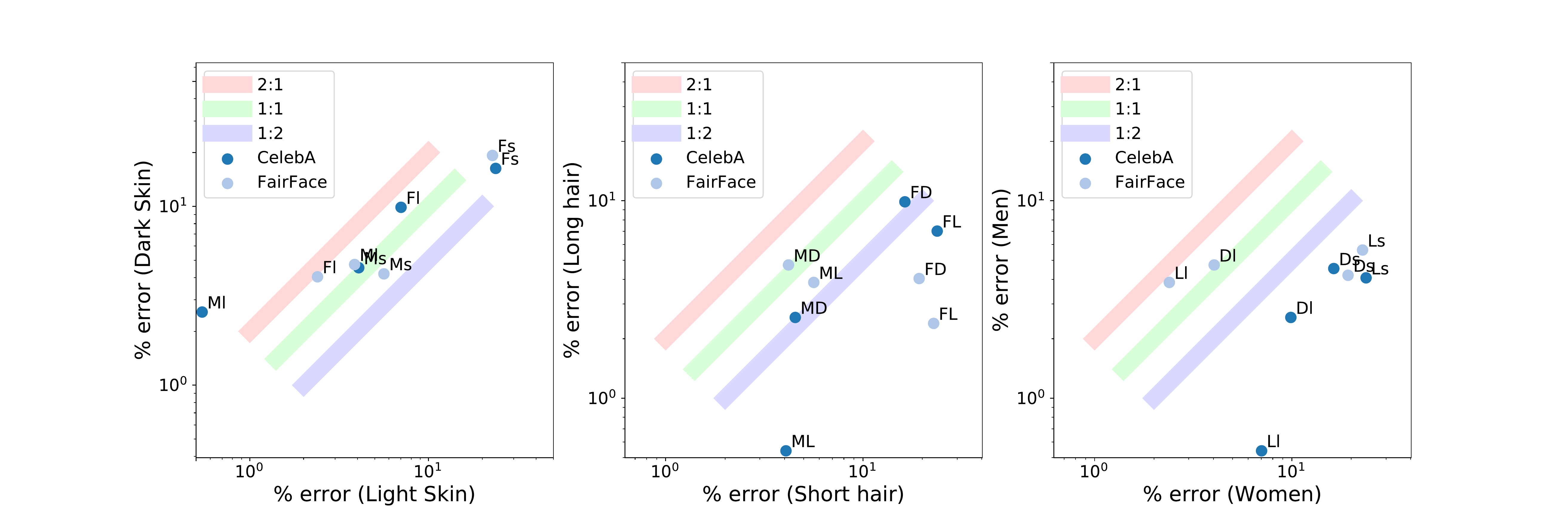}
\vspace{-0.15in}
\caption{\footnotesize \textbf{Scatter plots of error rates using data from Fig~\ref{fig:bar-plots} (transects).} Each dot compares the error rates of a pair of groups that differ by one attribute only (indicated in the label of the $x$ and $y$ axes).  The two letters near each dot indicate the shared attributes (`M/F' indicate male and female, `D/L' indicate dark and light skin, and `s/l' indicate short and long hair). Dots falling along the equal error line indicate that skin tone has little or no effect on error. In contrast, females and persons with short hair have higher error rates.}
\label{fig:scatter-plots}
\end{figure}

\subsection{Analysis of Bias}
We now analyze the performance of the classifiers on PPB and our transects. We verify that the classifiers exhibit similar error patterns to the commercial classifiers already evaluated on PPB~\cite{buolamwini2018gender}. Because PPB only consists of adults, we remove children and teenagers (age $< 0.4$ in the normalized $[0,1]$ scale) from our transects to make a more direct comparison, leaving us with 5335 total images.

Fig.~\ref{fig:gender-shades-plots} presents classification errors split by gender (M/F)  and skin color (L/D). We replicated the reported errors of the commercial classifiers in~\cite{buolamwini2018gender}, and report the errors of our classifiers on our in-house version of PPB. All classifiers perform significantly worse on dark-skinned females. Fig.~\ref{fig:bar-plots} and Fig.~\ref{fig:scatter-plots} present classification errors, stratified by gender/hair length/skin color combinations. We can make a number of broad-stroke, qualitative observations: 

\begin{itemize} 
\item The broad pattern of errors is similar across PPB and transects, with more errors on the left (females) than on the right (males).
\item Transect errors are either comparable or higher than in PPB, indicating that synthetic faces can be at least as challenging as real faces. Most significantly, errors are nonzero on males, which allows the study of relative difficulties when attributes are varied.
\item In PPB, there are few males with long hair and few females with short hair and light skin, making measurements unreliable for these categories. This is not a problem with transects, where faces are matched by attributes.
\item Transect errors are higher when hair is shorter for women. However, hair length has a negligible effect for males (see Fig.~\ref{fig:hair-and-beard} for a possible explanation). 
\item There is no consistent transect error pattern in skin tone: within homogeneous groups changing skin tone does not seem to affect the performance of either algorithm. For example, females with long hair see no significant difference in classification error between light vs. dark skin. Looking at PPB alone, we could not make this observation, since skin tone is so strongly correlated with hair length.
\end{itemize}

\begin{figure}[t]
\centering
\includegraphics[width=1\textwidth]{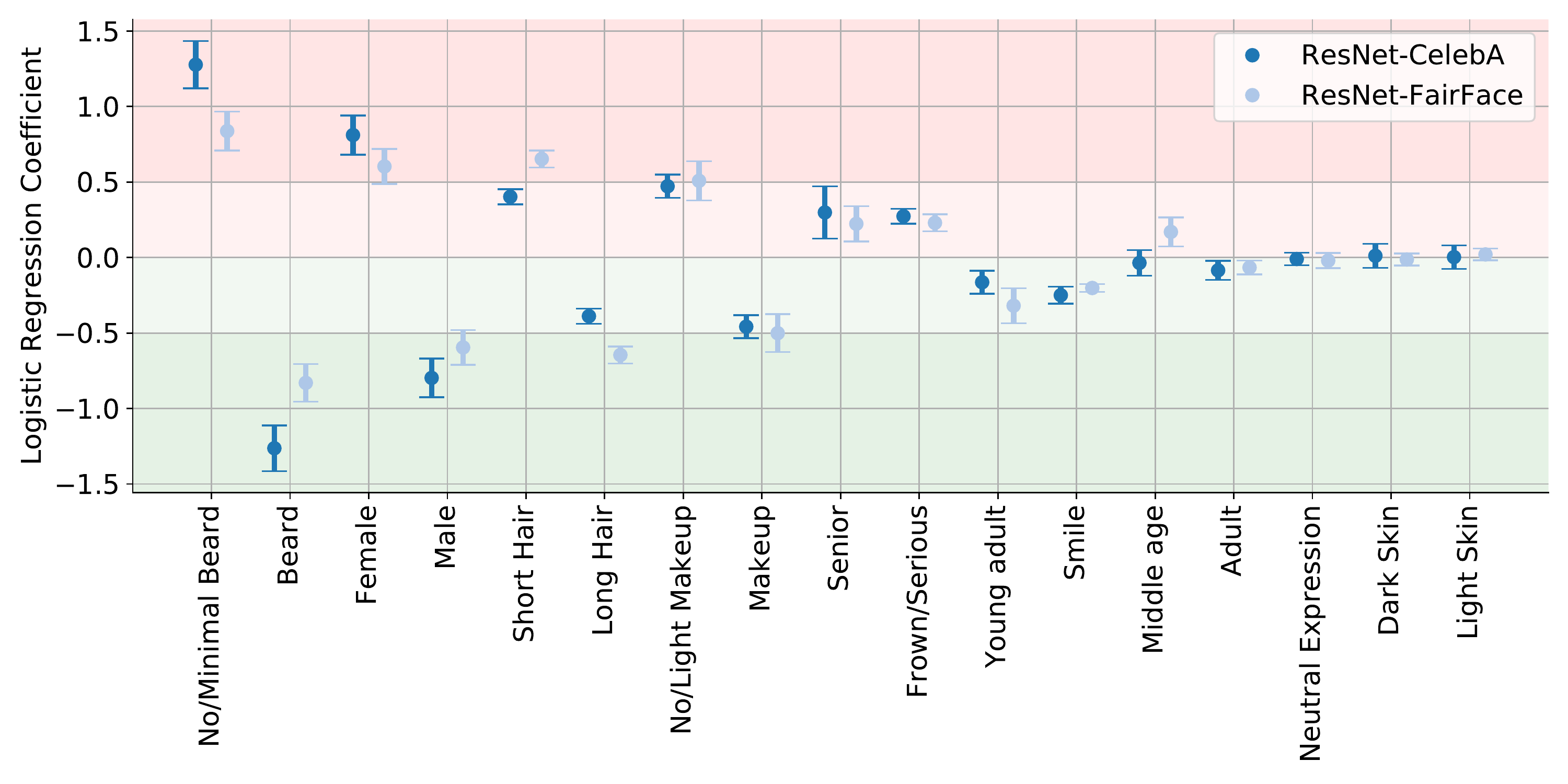}
\vspace{-1cm}
\caption{\footnotesize {\bf Logistic regression coefficient values.} The logistic regression model is trained to predict \emph{absolute errors} of the gender classifiers on our transect data given attributes as input. Coefficients represent the change in \emph{log odds} of the error for a change of 1 unit of each attribute. Larger coefficient magnitudes indicate more important variables, and positive(red)/negative(green) values correspond to variables that increase/decrease classifier error. Each attribute subgroup labeled on the $x$-axis is represented by a binary variable in the regression model, and we order attributes in this plot from large-to-small coefficient magnitudes. Error bars report standard deviations that were computed via bootstrapping 1000 times.}
\label{fig:regression-results}
\end{figure}

\subsection{Regression Analysis}
\label{sec:regression}
In order to obtain a quantitative assessment of effect (or lack thereof) of attributes on classifier error, we investigated further by calculating covariate-adjusted causal effects. For each gender classifier model we trained an $L2$-regularized logistic regression model to predict that classifier's error conditioned on all attributes. 

We discretized attributes into levels, and assigned a binary variable to each level. We used the same discretization for hair length (short vs. long hair), skin color (light vs. dark skin) and gender (female vs. male) used in our experiments thus far. We used two levels for beard (no/light beard vs. beard) and makeup (no/light makeup vs makeup), three for facial expression (serious/frown vs. neutral vs. smile), and the original semantic levels for age described in Fig.~\ref{fig:screenshots}. In all, this resulted in 17 input variables to our logistic regression model. We used scikit-learn's LogisticRegression function~\cite{scikit-learn}, and set the regularization parameter to 1. 

Fig.~\ref{fig:regression-results} presents coefficients for both logistic regression models. Recall that each coefficient represents the change in log odds of the classifier's error for a change of 1 unit of each covariate (see Sec.~\ref{sec:analysis}). Error bars depict standard deviations, obtained by bootstrapping the dataset 1000 times. A person's facial hair, gender, makeup, hair length and age all have significant effects on classification error, and skin color has a negligible effect. Our main experimental conclusion is that observational (PPB) and experimental (transects) methods are fundamentally at odds on the causes of algorithm bias in gender classification algorithms. Observational analysis on wild-collected PPB suggests that a combination of gender and skin tone are implicated, while our experimental method using synthetic transects suggests that other attributes are far more important than skin tone. 

\begin{figure}[t!]
\centering
{\bf Decision Threshold = 0.5}\\
\includegraphics[width=\textwidth]{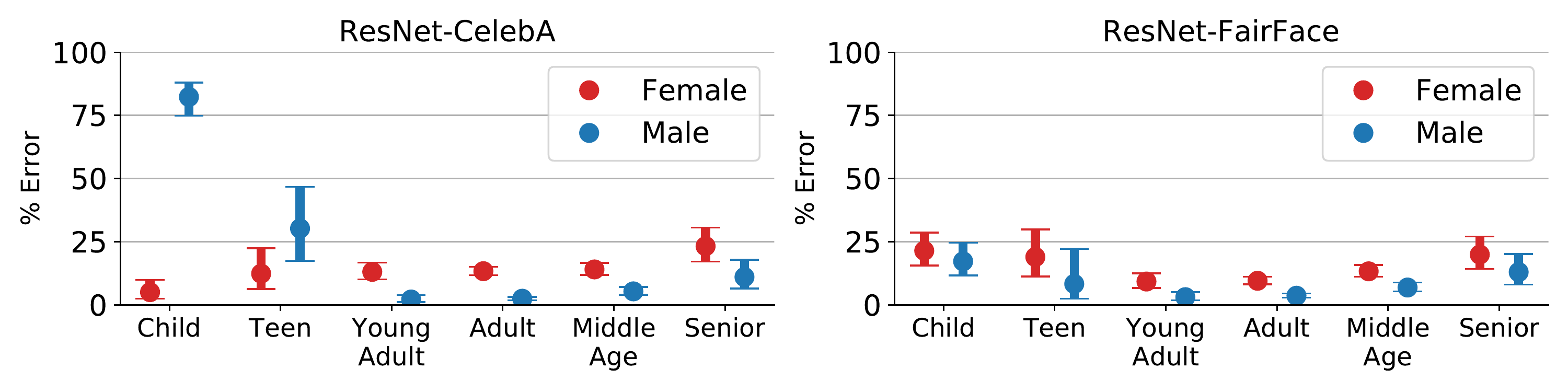}
{\bf Decision Threshold = 0.8}\\
\includegraphics[width=\textwidth]{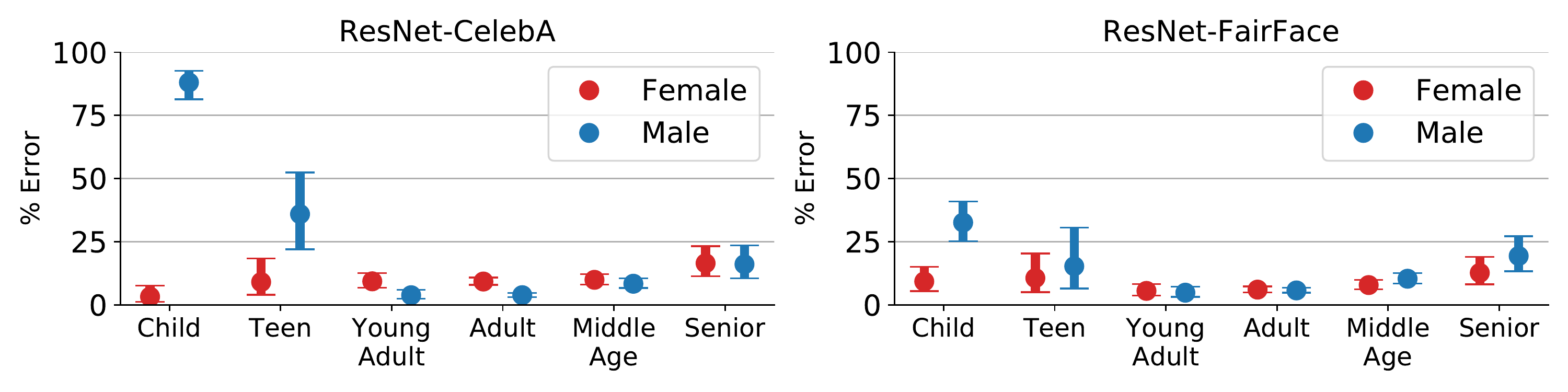}
\caption{\footnotesize{\bf Errors by gender and age group on our transect images.} The two top plots were obtained by using a decision threshold equal to 0.5, and show a prevalence of female errors. The bottom two plots were obtained with a threshold equal to 0.8, chosen to minimize overall error. There is a non-uniform influence of age on errors. Both models tend to have lower errors for young to middle-aged adults. The differences in errors between genders are fairly consistent for adults, but differ for children, teenagers and seniors, illustrating a combined age-gender bias in the algorithms.}
\label{fig:age-gender-joint}
\end{figure}

\begin{figure}[t!]
\includegraphics[width=\textwidth]{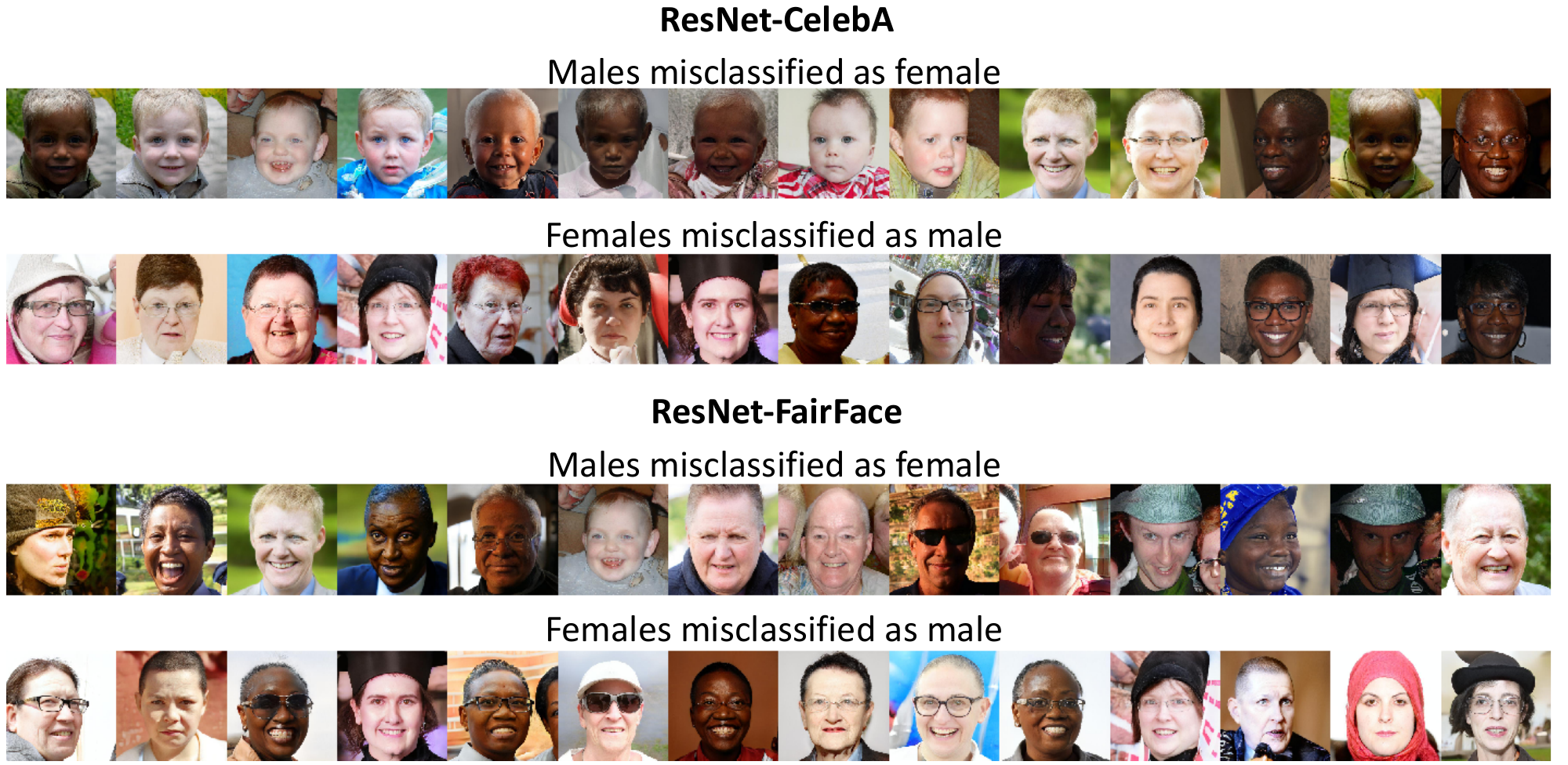}
\caption{\footnotesize{\bf Images with largest errors.} Synthetic faces on which the classifiers most deviated from the mean human annotations.}
\label{fig:misclassified}
\end{figure}

\subsubsection{Joint Effects of Attributes on Classification Error}
\label{sec:joint}
Our regression analysis makes a simplifying assumption that each covariate has an independent, linear effect on classification error. The independence assumption can be a poor one. For example, Fig.~\ref{fig:scatter-plots}-right shows that error rates vary across different intersectional groups of skin color and hair length in a way that is not simply a linear combination of each attribute.

This is also the reason we removed children and teenagers from our analysis, as these individuals tend to have different appearance characteristics from adults. Fig.~\ref{fig:age-gender-joint} illustrates this, by breaking down error rates by age and gender subpopulations for two classifier decision thresholds. The difference in error rates between the genders is fairly consistent for young adults to middle-aged individuals, but vary for children/teenagers and seniors. This demonstrates that age and gender have joint effects on errors. 

Fig.~\ref{fig:misclassified} shows faces from our synthesized transects on which the ResNet models were most incorrect. For each gender misclassification direction, we show faces on which the model predictions were farthest from the average human annotator response. ResNet-CelebA tends to heavily misclassify young male children/babies as female, in line with the quantitative result in Fig.~\ref{fig:age-gender-joint}.

\section{Discussion and Conclusions}
\label{sec-discussion-conclusions}
\subsection{Summary}
Our study leads us to three main conclusions. First, the experimental approach to measuring algorithmic bias in computer vision is feasible. Second, the experimental approach may yield quite different conclusions from traditional observational studies. Third, when analyzing algorithmic bias, a broad spectrum of attributes and attribute combinations should be considered besides the ones of immediate interest. We examine each in detail below. 

Our experimental approach is made possible by combining recent progress in image synthesis with detailed human annotations collected by crowdsourcing. Image synthesis, calibrated by human annotations, allows us to generate transects of matched samples, i.e., groups of images which differ only along attributes of interest. In contrast to previous attribute-specific methods~\cite{muthukumar2018understanding} {\em any} attribute may be explored, provided that it can be annotated by humans. By relying on human ground truth annotations, one does not need to rely on the synthesis method being perfect. 

\begin{figure}[t!]
\includegraphics[width=\textwidth]{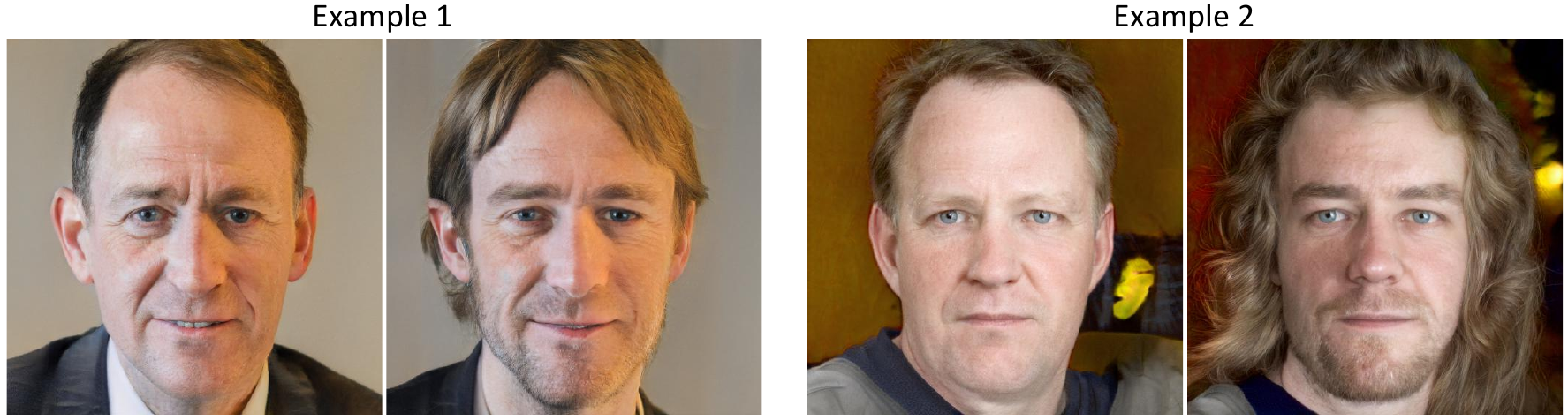}
\caption{\footnotesize{\bf Correlated attribute modifications.} We found that our method sometimes adds a beard to a male face when attempting to only modify hair length. This is an example of an imprecise intervention which can complicate downstream bias analyses. This bias may be due to the training data itself (men with long hair tend to have facial hair), or injected by the algorithm.}
\label{fig:hair-and-beard}
\end{figure}

\begin{figure}[t!]
\includegraphics[width=\textwidth]{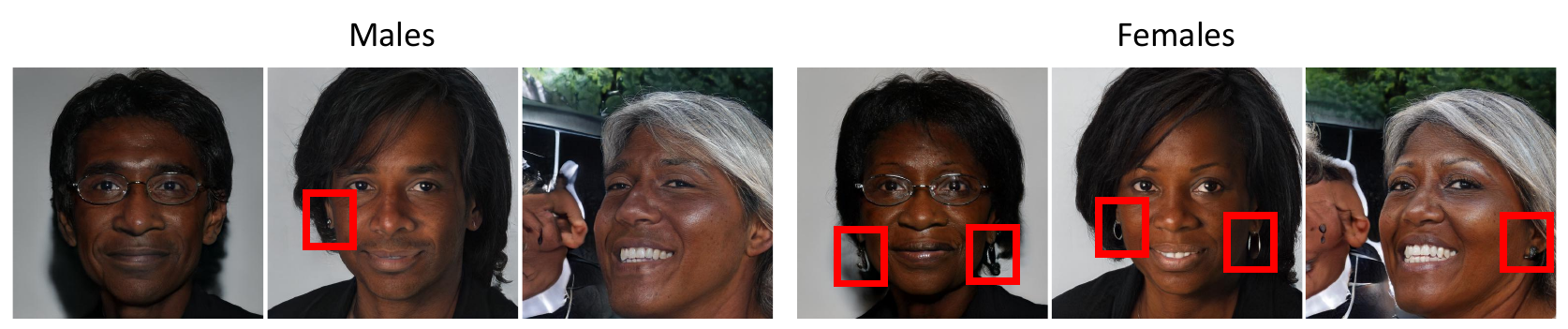}
\caption{\footnotesize{\bf Hidden confounders.} There is always the possibility of a hidden confounder lurking in a dataset. As an example, we found --- after already collecting annotations --- that our method tends to add earrings when transitioning from dark-skinned men to dark-skinned women, a cue that a gender classifier might use to perform disproportionately well on the latter group. Because we did not annotate this attribute, it is hidden to our analysis. Interestingly, one male in this image also has an earring; that earring becomes larger for his female counterpart.}
\label{fig:earrings}
\end{figure}

The experimental method and our synthesis-based experimental approach, offer a number of attractive properties and advantages over traditional observational methods:

\begin{enumerate}
\item {\bf Causal inferences on bias are possible.} Our method generates approximately matched samples across selected attributes, allowing for counterfactual analysis, e.g., {\em ``Would the algorithm have made the same mistake if the same person had had a different skin color?''} Observational image data are almost never matched.

\item {\bf Bias may be measured for underrepresented groups.} Image synthesis allows, to a great degree, uniform sampling of the space of attributes of interest --- gender, skin color and hair length in our experiments. This is very difficult to do when one relies on images that are sampled from natural distributions, which tend to be long-tailed and therefore where some groups are underrepresented.

\item {\bf Bias may be measured for intersectional groups.} Our method allows researchers to draw causal inferences across groups that are defined by specific attribute combinations. Single-attribute analysis may conceal biases affecting groups defined by the combination of multiple attributes~\cite{kearns2019ethical}. Some such combinations are often vastly undersampled in natural data.

\item {\bf Bias measurements are valid across different populations.} This is because the experimental method identifies causally linked attributes, independent of the prevalence of these attributes, i.e., the bias measurements are a property of the algorithm and not of the population on which it is used. By contrast, observational measurements do not generalize beyond the narrowly defined population where the data was collected. Furthermore, by combining appropriately the contribution of different attributes, one may predict the effects both of {\em disparate treatment}~\cite{mendez1980presumptions} and {\em disparate impact}~\cite{rutherglen1987disparate} on a specific population. 

\item {\bf Accurate bias measurements may be made quickly and inexpensively.} Image synthesis is fast and inexpensive, and crowdsourced image annotation is also relatively fast and affordable. By contrast, assembling large datasets of natural images is laborious and expensive -- it may take years and substantial investment, which may only be afforded by large organizations. Thus, synthetic data has the potential to democratize testing for bias. 


\item {\bf Ethical and legal concerns are greatly reduced.} Collecting face image datasets in the wild requires great care to respect the privacy and dignity of individuals, the rights of minors and other vulnerable groups, as well as copyright laws. By contrast, synthetic datasets are free from such risks because they do not depict real people. 
\end{enumerate}


The experimental analysis (transects) and traditional observational analysis (using PPB) diverged most significantly on the effect of skin color, which the observational study flagged as significant and the experimental method found to be not significant in determining algorithmic bias. The experimental method reveals a number of additional sources of bias: age, hair length and facial hair (Fig.~\ref{fig:regression-results}). The two methods agree on gender. Our analysis suggests that the difference between the conclusions of the two methods is likely due to the correlation of hair length, skin color and gender in PPB (see Fig.~\ref{fig:violin-gender} and Fig.~\ref{fig:violin-color}). Consequently, if one does not control for hair length, the classifiers' bias towards assigning gender on the basis of hair length is read as a bias concerning dark-skinned women. The triple correlation between hair length, gender and skin color had been noticed in a previous study~\cite{muthukumar2018understanding}.




The main reason for measuring algorithmic bias is to get rid of it. Error and bias measurements guide scientists and engineers towards effective corrective measures for improving the performance of their algorithms. It is instructive to view the different predictions of the two methods through this lens. The correlational study based on PPB (Fig.~\ref{fig:gender-shades-plots}) may suggest that, in order to reduce biases in our classifiers, more images of dark-skinned women should be added to their training sets. The experimental method leads engineers in a different direction. First, more training images of long-haired men and short-haired women of all races are needed. Second, correcting age bias requires more training images in the child-teen and, possibly, senior age groups.

Finally, a lesson from our study is that it is important to consider a rich number of attributes and attribute combinations, besides the one(s) of immediate interest. This is for two reasons. First, unobserved confounders can have strong effects and need to be included in the analysis. Second, the combined effect of attributes can be strongly nonlinear (see the interaction of age and gender in Fig.~\ref{fig:age-gender-joint}), and therefore an intersectional analysis~\cite{kearns2017preventing,buolamwini2018gender} is necessary. Selecting attributes or attribute combinations is as much of an art as a science, and therefore one has to rely on good judgment and on a healthy multidisciplinary debate to progressively reveal missing ones.

\subsection{Limitations and Future Work}
While the advantages of the experimental method are clear, our proposed method does not exempt researchers from exercising attention and good judgment. In particular, while our method greatly reduces unwanted correlations with annotated variables, it does not eliminate them completely, nor does it account for hidden confounders~\cite{vanderweele2013definition}, and one will need to keep a sharp eye out for both. As an example of the first, we found that our method often adds facial hair to male faces when increasing hair length (see Fig.~\ref{fig:hair-and-beard}). This is likely a reason for why our classifiers did not have higher error rates for males with longer hair (see Fig.~\ref{fig:bar-plots}). As an example of the second, we found that our method tends to synthesize earrings when modifying a dark-skinned face to look female (see Fig.~\ref{fig:earrings}). Depending on culture, earrings may or may not be relevant to the definition of gender. If this is an unwanted correlation, one ought to add earrings to the annotation pipeline so that it may be ``orthogonalized away'' by the synthesis method. Scientists building an industry-grade system for measuring face analysis bias will want to consider including a more exhaustive set of factors. A significant advantage of an approach that is based on synthetic images and human annotation is thus the following: {\em as soon as one residual correlation is discovered it may be systematically annotated, compensated for in the analysis, and mitigated in the synthesis.}

A number of refinements in face synthesis will make our experimental method more practical and powerful. First, many of the faces we generated contained visible artifacts (see Fig.~\ref{fig:annotation-stats}), which we eliminated by human annotation -- even subtle artifacts can affect classifier outputs, as revealed by the literature on adversarial examples~\cite{szegedy2013intriguing}. Second, we do not yet have tools to estimate the sets of physiognomies and attribute combinations that can and cannot be produced by a given generator. Current GANs are known to have difficulties in generating data outside of their training distributions. Third, we observed a bias of StyleGAN2 towards generating Caucasian faces when sampling from its latent distribution. While our method can compensate for biases through carefully oriented traversals calibrated by human annotations, it would be clearly better to start from unbiased synthesis methods. We are hopeful that these shortcomings will be incrementally resolved by a combination of training sets with increased diversity of attributes like ethnicity, gender, personal style, and age, as well as better models.

Our first-order technique for controlling synthesis can also be improved. A better understanding of the geometry of face space will hopefully yield more accurate global coordinate systems. These, in turn, will help reduce residual biases in synthetic transects, which we currently mitigate by having transects annotated by hand.

Finally, extending our method beyond gender classification to more complex tasks, such as face recognition, is not straightforward in practice and will require further study.

\section*{Acknowledgments}
A number of colleagues kindly read draft versions of this manuscript, providing references, insightful comments and valuable criticisms. We are especially grateful to Frederick Eberhardt, Bill Freeman, Lei Jin, Michael Kearns, R. Manmatha, Tristan McKinney, Sendhil Mullainathan, and Chandan Singh.

\newpage
\bibliographystyle{splncs04}
\bibliography{egbib}

\end{document}